\documentclass[lettersize,journal]{IEEEtran}
\usepackage[switch]{lineno}
\usepackage{array}
\usepackage{stfloats}
\usepackage{verbatim}
\usepackage{threeparttable}
\usepackage{cite}
\usepackage{amsmath,amssymb,amsfonts}
\usepackage{algorithmic}
\usepackage{graphicx}
\usepackage{subfigure}
\usepackage{textcomp}
\usepackage{xcolor}
\usepackage{colortbl}
\usepackage{float}
\usepackage{booktabs}
\definecolor{awesome}{rgb}{0.98, 0.66, 0.68}
\usepackage{multirow}
\usepackage{etoolbox}
\usepackage{setspace}
\usepackage{etoolbox}
\usepackage{algorithmic}
\usepackage{algorithm}
\usepackage{ragged2e}
\usepackage[edges]{forest}
\usepackage{tikz}
\usetikzlibrary{shapes,positioning}
\usepackage[colorlinks=true, linkcolor=blue, citecolor=blue, urlcolor=magenta]{hyperref}

\tikzset{
  ft_root/.style={
    rectangle,
    draw=blue!80,
    fill=blue!3,
    thick,
    rounded corners,
    align=center,
    text width=2cm,
    font=\footnotesize
  },
  ft_sec/.style={
    rectangle,
    draw=blue!80,
    fill=blue!18,
    rounded corners,
    align=center,
    text width=2cm,
    font=\footnotesize
  },
  ft_subsec/.style={
    rectangle,
    draw=blue!80,
    fill=blue!12,      
    rounded corners,
    align=center,
    text width=2.5cm,
    font=\scriptsize
  },
  ft_subsubsec/.style={
    rectangle,
    draw=blue!80,
    fill=blue!6,
    rounded corners,
    align=left,
    text width=9.5cm,
    font=\scriptsize
  },
  edge/.style={
    draw=blue!80,
    -latex,
    line width=0.3mm
  },
  level distance=8mm,
  sibling distance=5mm
}

\def\BibTeX{{\rm B\kern-.05em{\sc i\kern-.025em b}\kern-.08em
    T\kern-.1667em\lower.7ex\hbox{E}\kern-.125emX}}
\usepackage{balance}
\begin{document}

\title{Parallels Between VLA Model Post-Training \\ and Human Motor Learning: Progress, \\Challenges, and Trends \\

\thanks{Tian-Yu Xiang, Ao-Qun Jin, Xiao-Hu Zhou, Mei-Jiang Gui, Xiao-Liang Xie, Shi-Qi Liu, Shuang-Yi Wang, Sheng-Bin Duan, Fu-Chao Xie, Wen-Kai Wang, Si-Chang Wang, and Ling-Yun Li are with the State Key Laboratory of Multimodal Artificial Intelligence Systems, Institute of Automation, Chinese Academy of Sciences, Beijing 100190, China, and also with the School of Artificial Intelligence, University of Chinese Academy of Sciences, Beijing 100049, China (e-mail:
xiangtianyu2021@ia.ac.cn; xiaohu.zhou@ia.ac.cn).}
\thanks{Tian Tu is with The Grainger College of Engineering, University of Illinois Urbana-Champaign}
\thanks{Zeng-Guang Hou is with State Key Laboratory of Multimodal Artificial Intelligence Systems, Institute of Automation, Chinese Academy of Sciences, Beijing 100190, China, also with the CAS Center for Excellence in Brain Science and Intelligence Technology, Beijing 100190, China, also with the School of Artificial Intelligence, University of Chinese Academy of Sciences, Beijing 100049, China, and also with the Joint Laboratory of Intelligence Science and Technology, Institute of Systems Engineering, Macau University of Science and Technology, Taipa, Macao, China (e-mail: zengguang.hou@ia.ac.cn).}
\thanks{$^{\dagger}$ These authors contributed equally: Tian-Yu Xiang, Ao-Qun Jin}
\thanks{$*$ Corresponding author: Xiao-Hu Zhou, Zeng-Guang Hou.}
}
\author{\IEEEauthorblockN{Tian-Yu Xiang$^{\dagger}$, Ao-Qun Jin$^{\dagger}$, Xiao-Hu Zhou$^{*}$, \emph{Member, IEEE}, Mei-Jiang Gui, Xiao-Liang Xie, \emph{Member, IEEE}, Shi-Qi Liu, Shuang-Yi Wang, Sheng-Bin Duan, Fu-Chao Xie, Wen-Kai Wang, Si-Cheng Wang, Ling-Yun Li, \\Tian Tu, Zeng-Guang Hou$^{*}$, \emph{Fellow, IEEE}}
}

\markboth{Journal of \LaTeX\ Class Files,~Vol.~xx, No.~x, xx~xxxx}%
{How to Use the IEEEtran \LaTeX \ Templates}\maketitle

\begin{abstract}

Vision-language-action (VLA) models extend vision-language models (VLM) by integrating action generation modules for robotic manipulation. Leveraging the strengths of VLM in vision perception and instruction understanding, VLA models exhibit promising generalization across diverse manipulation tasks. However, applications demanding high precision and accuracy reveal performance gaps without further adaptation. Evidence from multiple domains highlights the critical role of post-training to align foundational models with downstream applications, spurring extensive research on post-training VLA models. VLA model post-training aims to enhance an embodiment’s ability to interact with the environment for the specified tasks. This perspective aligns with Newell's constraints-led theory of skill acquisition, which posits that motor behavior arises from interactions among task, environmental, and organismic (embodiment) constraints. Accordingly, this survey structures post-training methods into four categories: (i) enhancing environmental perception, (ii) improving embodiment awareness, (iii) deepening task comprehension, and (iv) multi-component integration. Experimental results on standard benchmarks are synthesized to distill actionable guidelines. Finally, open challenges and emerging trends are outlined, relating insights from human learning to prospective methods for VLA post-training. This work delivers both a comprehensive overview of current VLA model post-training methods from a human motor learning perspective and practical insights for VLA model development. Project website: \href{https://github.com/AoqunJin/Awesome-VLA-Post-Training}{Awesome-VLA-Post-Training}.

\emph{Impact Statement}—The VLA model, a type of robotic manipulation foundation model, enables robotic systems with the abilities of task understanding and environmental perception. However, current implementations require an adaptation process—termed “post-training”—analogous to human motor skill learning before practical deployment. This review systematically examines post-training strategies through the lens of human motor learning (based on Newell's constraints-led theory), focusing on three key components: environments, embodiments, and tasks. By synthesizing insights from neuroscience and advanced AI, this study provides a novel perspective for understanding how to enhance a pre-trained VLA model. The proposed perspective bridges neuroscience and robotics, aligning with the motivation of NeuroAI. It also offers guidelines to post-train VLA manipulation models, lowering barriers for future research.

\end{abstract}


\section{Introduction}

Building on the success of foundation models in computer vision (CV) and natural language processing (NLP)~\cite{awais2025foundation, wang2025comprehensive}, recent research has extended the foundation model paradigm to robotics. Vision-language-action (VLA) models have emerged as a leading class of robotic foundation models~\cite{brohan2023rt}, integrating the perceptual capabilities of vision-language models (VLM) with action generation modules. These models leverage visual perception to capture environmental context and language understanding to interpret task instructions. Action sequences are thus generated by conditioning on the combined vision-language representations.

\begin{figure}[tb]
   \centering
     \includegraphics[width=3.5in]{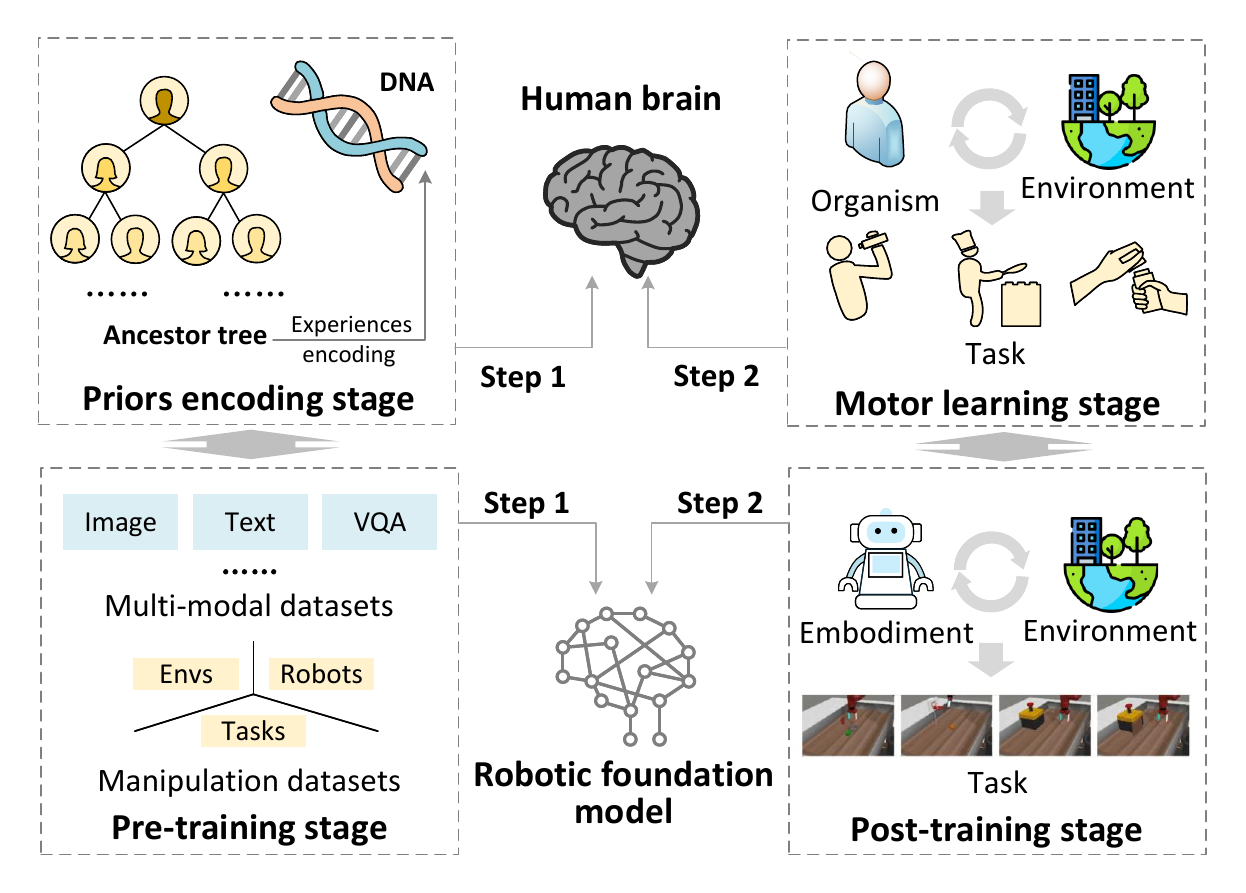}
    \caption{Parallels between human motor learning and robotic foundation model training. Both processes begin with encoding broad priors from general experiences, followed by task-specific adaptation within target environments.}
    \label{fig1}
\end{figure}

VLA-based policies differ from conventional robotic policies in their reliance on extensive pre-training~\cite{o2024open}. VLA models are typically initialized from a pre-trained vision-language encoder—either a VLM or separate vision and language encoders model—and subsequently trained on large-scale heterogeneous datasets encompassing diverse manipulation tasks, multiple robotic embodiments, varying environments, and multimodal data from non-robotic domains. This paradigm exposes VLA models to broad scenarios, enhances manipulation comprehension, and formulates generalized policies for multi-task robotic applications.

Benefiting from extensive pre-training, recent state-of-the-art (SOTA) VLA models can perform zero-shot tasks~\cite{intelligence2025pi_, kimopenvla}. However, in contrast to the remarkable success of foundation models in NLP~\cite{achiam2023gpt}, VLA models display lower out-of-the-box performance and stability, which hinders their readiness for real-world deployment. This gap reflects the challenges inherent in training foundation models for robotic manipulation:

\begin{itemize}
    \item Limited datasets for open environments: Despite recent growth, robotic manipulation datasets remain small compared to those in NLP. For example, the largest open-source robotic manipulation dataset~\cite{o2024open} contains approximately 2.5 million demonstrations, even fewer than pre–large-language-model-era NLP corpora, such as GPT-2’s training set of 8 million web pages~\cite{radford2019language}.
    \item Heterogeneous embodiments and execution styles: Robotic systems exhibit varied forward and inverse dynamics~\cite{kirschner2025categorizing}. Furthermore, the execution strategies are infinite for a fixed task under identical initial conditions. By comparison, once semantic content is specified, linguistic expression is comparatively constrained.
    \item Complex tasks with abstract rules: Manipulation tasks rely on highly abstract rules that are difficult to generalize. In contrast, linguistic structures, though complex, can be systematically described~\cite{wardhaugh2002understanding}.
\end{itemize}

The challenges in environments, embodiments, and tasks underscore the need to adapt robotic foundation models for downstream tasks. This perspective aligns with Newell's constraints-led theory of human skill learning~\cite{newell1986constraints}, which holds that motor behavior emerges from interactions among task, environmental, and organismic (embodiment) constraints. Recent successes in adaptation for NLP~\cite{shah2023creation,singhal2023large} and CV~\cite{wu2025medical,chen2024ma} motivate the investigation of adaptation approaches for VLA models. Adaptation is especially critical for VLA models, as their generalization remains lower than that of language and vision models~\cite{intelligence2025pi05,kimopenvla}.

The adaptation process, often termed ``post-training'' or ``fine-tuning'', differs from the pre-training stage, which leverages large-scale, heterogeneous datasets to enhance generalization~\cite{quinn2019dive}. Post-training targets specific downstream applications with relatively small, specialized datasets, shifting the model’s objective from broad generalization to optimized performance within an application domain. In robotic manipulation, adaptation denotes the optimization of a pre-trained VLA model for a particular embodiment, improving performance on one or more manipulation tasks in target environments. This process mirrors human skill acquisition: innate priors encoded in the genome guide initial learning stages~\cite{shadmehr2004computational,medawar1983aristotle}, while subsequent skill refinement through interaction and practice corresponds to post-training~\cite{newell1986constraints}.

The emergence of VLA models and other foundation models in robotic learning has spurred numerous literature reviews~\cite{firoozi2023foundation, jaquier2025transfer, gao2025taxonomy, ma2024survey, sapkota2025vision, xiao2025robot, hu2023toward, kawaharazuka2024real}. These reviews provide diverse perspectives on the era of robotic foundation models. However, despite increasing attention to VLA model adaptation~\cite{ma2024survey, sapkota2025vision}, a comprehensive review of VLA model post-training techniques is still limited. This review addresses that gap by focusing on VLA model adaptation, offering a structured taxonomy of post-training methods through the lens of human motor learning, and highlighting key challenges for future research.

The contributions can be summarized as follows:
\begin{itemize}
    \item This study provides a comprehensive review of the VLA model post-training, examining it through the lens of human motor learning.
   \item Paralleled with Newell's constraints-led theory of human skill learning~\cite{newell1986constraints}, a taxonomy of post-training VLA methods is organized around environments, embodiments, and tasks.
    \item This review presents current insights on post-training VLA models, delineating key challenges and future research directions in the application of VLA models.
\end{itemize}

The remainder of this paper is organized as follows. Section II reviews background on VLA models and human motor learning. Section III details the methodology used in this review. Section IV introduces the proposed taxonomy and related studies. Section V presents comparative evaluations of post-training methods on standard benchmarks. Section VI analyzes key challenges and relates human strategies for addressing them. Section VII concludes this review.

\section{Background}

\subsection{Explosion of Robot Manipulation Datasets}

\begin{figure}[tb]
  \centering
  \includegraphics[width=3.5in]{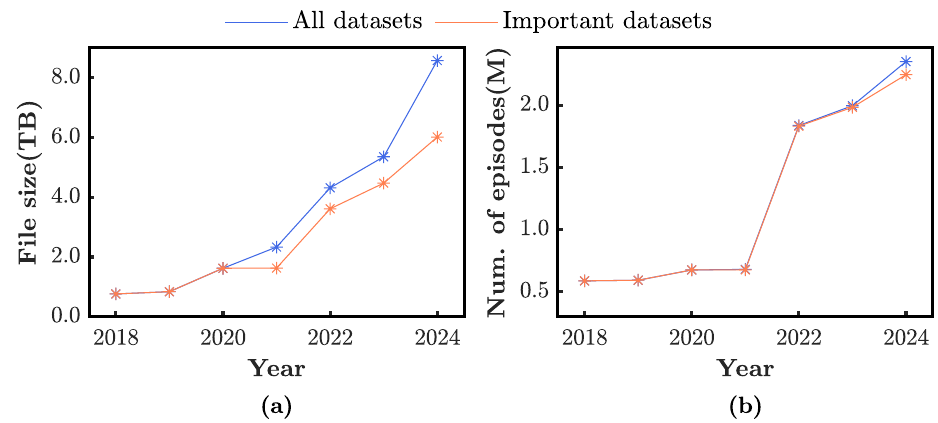}
  \caption{Evolution of manipulation dataset scale in the Open X-Embodiment collection~\cite{o2024open}. Important datasets are defined as those with over 50 citations or published after 2023. (a) File size in terabytes (TB). (b) Number (Num.) of episodes in millions (M).}
  \label{datascale}
\end{figure}

Similar to trends in other domains, manipulation datasets for robots have undergone rapid expansion. The Open X-Embodiment dataset~\cite{o2024open} aggregates 58 existing robotic manipulation datasets spanning diverse embodiments. As shown in Fig.~\ref{datascale}, it captures the cumulative scale of robot manipulation data collected between 2018 and 2024.

Over this period, total file size grew by more than an order of magnitude, while episode count increased by nearly fourfold. The faster growth in file size relative to episode count reflects improvements in dataset richness, with additional modalities and finer-grained demonstrations per episode. Restricting the analysis to 42 ``important” datasets—defined as those cited more than 50 times before June 1, 2025, or published after 2023—yields comparable total file size and episode counts, indicating that the influential datasets are inherently large.

In addition to real-world manipulation datasets, several high-fidelity simulators~\cite{hussain2020unity, coumans2015bullet, todorov2012mujoco, makoviychuk2isaac} and task-specific simulated environments~\cite{yu2020meta, liu2023libero, mees2022calvin} have emerged. By employing rule-based scripts or reinforcement learning (RL)–based policies as near-optimal experts, these platforms reduce the cost of generating manipulation datasets. Preliminary experiments indicate that a data-scaling law also applies to imitation learning~\cite{lin2024data, tuylsscaling}. The growing availability of large-scale manipulation datasets, together with advanced simulators, accelerates the development of robotic manipulation foundation models.

\subsection{General Structure of VLA Models}

\begin{figure}[tb]
  \centering
  \includegraphics[width=3.5in]{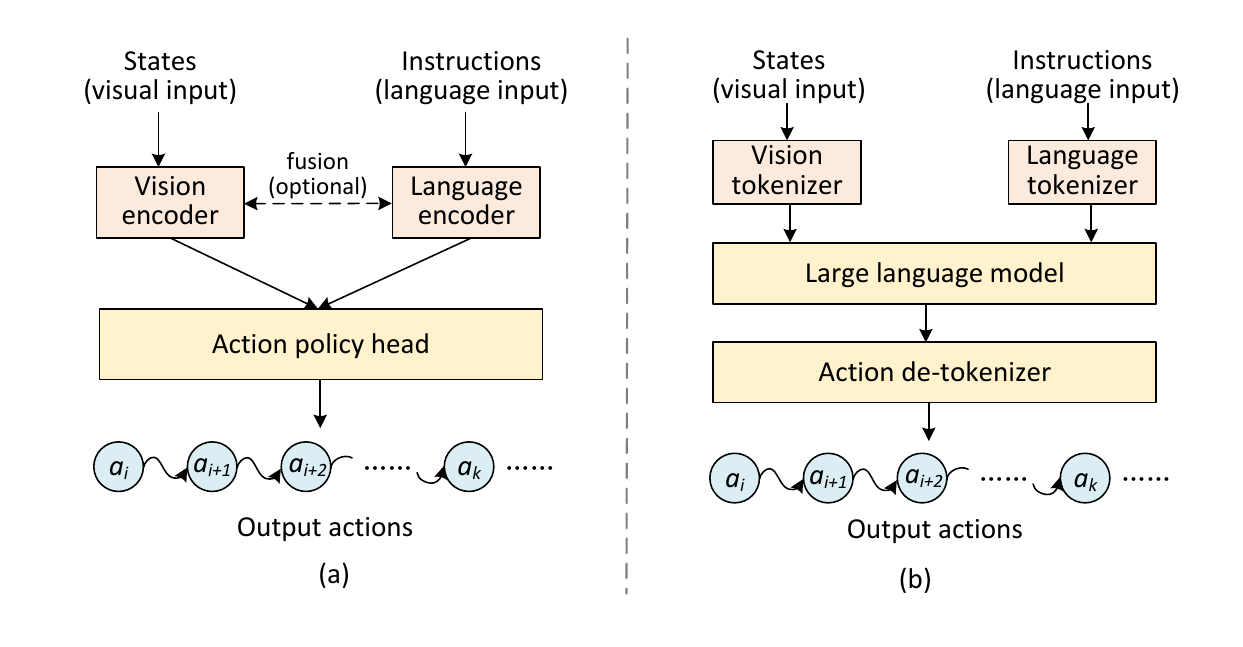}
  \caption{Prevalent VLA model architectures. (a) Visual and linguistic inputs are encoded separately; the resulting embeddings are passed to an action head for decoding. (b) Inputs are tokenized and processed by a pre-trained large language model (LLM) or multimodal foundation model; the output tokens are then decoded into actions by a detokenizer that serves as the action head.}
  \label{fig2}
\end{figure}

The basic architecture of VLA models is shown in Fig.~\ref{fig2}. These architectures encode visual and linguistic inputs and decode the resulting representations into action sequences~\cite{brohan2023rt, kimopenvla, intelligence2025pi_, shukor2025smolvla}. Variants extend this design by incorporating additional modalities, such as proprioceptive data (e.g., joint angles and positions)~\cite{wu2023unleashing, team2024octo, bjorck2025gr00t}. Despite these extensions, vision-language understanding remains central to robotic systems, and they are classified as VLA models.

The vision encoder in Fig.~\ref{fig2} (a) (or vision tokenizer in Fig.~\ref{fig2} (b)) processes vision context using a pre-trained model such as DINOv2~\cite{oquab2024dinov2} with backbone architectures including ResNet~\cite{he2016deep}, EfficientNet~\cite{tan2019efficientnet}, and the Vision Transformer~\cite{dosovitskiyimage}. The language encoder in Fig.~\ref{fig2} (a) (or large language model in Fig.~\ref{fig2} (b)) parses textual instructions. Current implementations range from SOTA foundation models~\cite{team2024gemma, touvron2023llama} to earlier encoders such as the CLIP text backbone~\cite{radford2021learning}. Owing to large-scale pre-training, these modules provide robust feature extraction.

Vision and language representations are fused into joint embeddings before action decoding. Fusion can occur during feature extraction (e.g., FiLM~\cite{perez2018film}) or post-extraction via feature concatenation or cross-attention~\cite{du2023learning}. Action policy heads vary by implementation: multilayer perceptrons are common, while Gaussian mixture models~\cite{reynolds2009gaussian} and diffusion-based heads~\cite{chi2023diffusion} can be employed to represent multimodal manipulation.

Foundation VLA models following this general structure have shown strong potential on robotic manipulation tasks~\cite{brohan2023rt, kimopenvla, intelligence2025pi_, intelligence2025pi05, wu2023unleashing, cheang2024gr, team2024octo, liurdt, li2024cogact, liu2024robomamba}. A notable trend is the use of auxiliary datasets in addition to pure manipulation data~\cite{kimopenvla, li2024cogact, team2024octo}, including visual question answering~\cite{brohan2023rt, liu2024robomamba}, internet videos~\cite{wu2023unleashing, cheang2024gr, intelligence2025pi_, intelligence2025pi05}, and the foundation vision tasks~\cite{team2025gemini}. This large-scale, heterogeneous training, together with the generalization capacity of vision-language encoders, yields broader generalization than traditional robot policies, with demonstrations on single-arm~\cite{deng2025graspvla}, bi-arm~\cite{liurdt}, and humanoid platforms~\cite{bjorck2025gr00t}. Nevertheless, out-of-the-box performance and reliability of the foundation VLA model remain limited, underscoring the need for post-training.

\subsection{Post-training and Fine-tuning the VLA Model}

Although the term ``fine-tuning" predates the foundation model era~\cite{pratt1991direct,hinton2006fast}, its adoption surged after Devlin et al. introduced BERT~\cite{devlin2019bert}. Fine-tuning denotes a transfer-learning technique wherein the parameters of a pre-trained neural network are further optimized on task-specific data~\cite{quinn2019dive}. The term ``post-training'' emerged alongside the rise of large-scale pre-training for foundation models and encompasses broader objectives than fine-tuning, including preference alignment and safety assurance~\cite{tie2025survey,quinn2019dive}.

In VLA models, evaluation metrics extend beyond task success to include factors such as stability and safety. To maintain terminological consistency, this paper adopts ``post-training'' to denote this comprehensive adaptation process. Post‐training for VLA models aims to enhance the model's performance, such as success rate and safety, for a given robotic embodiment operating in specific environments and task sets. This process inherently involves adaptation across three dimensions: environments, embodiments, and tasks. Accordingly, post‐training strategies can be classified as follows:
\begin{itemize}
    \item \textbf{Environments:} Enhancing perception capabilities in diverse operational contexts.
    \item \textbf{Embodiments:} Precisely modeling the dynamics of the agent or robotic embodiment.
    \item \textbf{Tasks:} Incorporating prior knowledge of manipulation.
\end{itemize}

These components may interact during adaptation. For example, full‐parameter supervised fine‐tuning establishes a direct mapping from environmental states to actions, thereby concurrently enhancing environmental perception and embedding task‐specific priors. Nonetheless, the proposed taxonomy offers two major advantages: (1) it mirrors key components of human motor learning processes, and (2) it enables enhancement of individual dimensions while revealing limitations in VLA models, providing practical guidance for deploying pre‐trained VLA models in downstream applications.

\subsection{Human Motor Learning and Newell's Constraints-led View}

One of the key motivations for studying human motor learning stems from its relevance to robotic engineering~\cite{shadmehr2004computational}. On one hand, advanced robotic control theories inform neuroscientific investigations of underlying neural mechanisms; on the other hand, insights from human motor learning guide the development of robotic learning algorithms. Thus, research in robotics and human neuroscience forms a synergistic, bidirectional evolution~\cite{qiao2023brain, prescott2023understanding}, aligning with the NeuroAI initiative’s aim of reciprocal advancement between artificial intelligence and neuroscience~\cite{zador2023catalyzing}.

The development of human motor skills can be illustrated by the well-established constraints-led framework proposed by Newell~\cite{newell1986constraints}, which conceives motor development as the emergence of coordination from interacting constraints. The framework classifies constraints into environmental, organismic, and task factors, and posits that the ``optimal'' coordination pattern for an action arises from their interaction. This perspective aligns with robot skill learning and provides a lens for analyzing post-training strategies:
\begin{itemize}
\item \textbf{Environmental constraints:} In human learning, environmental constraints are independent of the learner and are relatively time-invariant during training (e.g., scene layout, object properties, lighting). In robot learning, the environment is likewise external to the agent and typically fixed within an experimental session; post-training therefore emphasizes improving perception of the environment (visual, semantic, and depth cues) and their integration.
\item \textbf{Organismic constraints:} For humans, organismic constraints refer to individual-specific properties (e.g., body mass, limb lengths) that directly shape action-level outputs. For robots, these constraints are induced by the embodiment morphology, kinematics, actuation, and sensor placement. These factors directly affect moments of inertia and the forward and inverse models of manipulation.
\item \textbf{Task constraints:} Task constraints specify the goals associated with the desired outcome, which are shared between humans and robots. In robot learning, analogous task specifications guide skill acquisition; strengthening task comprehension refines the relevant constraints and facilitates the learning of manipulation skills.
\end{itemize}

The analogy between human motor learning and the pre-/post-training paradigm of VLA models is illustrated in Fig.~\ref{fig1}. Innate, ancestral priors encoded in the genome~\cite{shadmehr2004computational,medawar1983aristotle} constitute a biological ``pre-training'' stage. Task-specific skill acquisition then proceeds through interaction and practice, mirroring model post-training (Fig.~\ref{fig1}). At the task-learning stage, following Newell's constraints-led framework~\cite{newell1986constraints}, learning can be organized along three dimensions: environmental perception, embodiment awareness (e.g., proprioception), and task comprehension, which together progressively refine motor performance under given environmental conditions.

Recent neuroimaging studies comparing human perceptual and cognitive processes with AI algorithms have yielded valuable insights~\cite{conwell2024large, mischler2024contextual, cross2021using}. Rather than detailing neuron mechanisms, this survey emphasizes findings in human learning that parallel post-training strategies in VLA models. Although subsequent sections briefly discuss selected micro-level neural mechanisms, the primary emphasis is on the similarities between human motor learning and the VLA model post-training. By summarizing VLA post-training methods through the lens of human motor learning, the survey aims to inform and motivate advances in robotic learning algorithms.

\subsection{Related Reviews}

Recent advances in robotic manipulation have spurred numerous surveys on VLA models~\cite{ma2024survey, sapkota2025vision, din2025vision} and robotic foundation models~\cite{firoozi2023foundation, jaquier2025transfer, gao2025taxonomy, xiao2025robot, hu2023toward, kawaharazuka2024real}. Several provide broad overviews that synthesize the evolution, challenges, and future directions for VLA models~\cite{ma2024survey, sapkota2025vision, din2025vision} and general robotic foundation models~\cite{firoozi2023foundation, xiao2025robot, hu2023toward}. Other reviews target specific topics, including a taxonomy for evaluating robotic generalization~\cite{gao2025taxonomy}, transfer learning strategies~\cite{jaquier2025transfer}, and real-world applications~\cite{kawaharazuka2024real}. These reviews provide diverse perspectives on foundation models in robotic manipulation.

Despite the breadth of related work and the significance of post-training to downstream VLA applications, a comprehensive synthesis of post-training techniques specific to VLA is currently limited. Existing summaries of post-training primarily focus on foundation language models~\cite{zheng2025learning, hanparameter, tie2025survey}. A closely related survey~\cite{jaquier2025transfer} covers transfer learning in robotics, but transfer learning constitutes only one subset of the broader post-training spectrum. In contrast, this study confines its scope to manipulation-oriented VLA models and offers a structured taxonomy of VLA-specific post-training methods through the lens of human motor learning, highlighting key challenges for future research.

\begin{figure*}[!htb]
  \centering
  \begin{forest}
  for tree={
    font=\footnotesize,
    grow'=0,
    forked edges,
    anchor=west,
    s sep=1.8mm,
    l sep=4mm,
    edge={-latex, line width=0.3mm},
  }
    [\textbf{Post-training VLA Models}, ft_root
      [\textbf{Enhancing Environmental Perception (Sec. \ref{Sec:Enhancing Environmental Perception})}, ft_sec
        [Affordance-Guided\\Learning (Sec. \ref{Sec:Affordance-Guided Learning}), ft_subsec
            [\textbf{Definition:} Inject object-centric affordances (i.e. where and how to act) into the VLA to couple perception with action opportunities.,ft_subsubsec];
            [\textbf{Examples:} ControlVLA~\cite{li2025controlvla};
            CoA~\cite{li2024improving};
            A3VLM~\cite{huanga3vlm};
            KAGI~\cite{lee2024affordance};
            RoboPoint~\cite{yuanrobopoint};
            OmniManip~\cite{pan2025omnimanip};
            RT-Affordance~\cite{nasirianyrt};
            A0~\cite{xu2025a0};
            RoboCodeX~\cite{mu2024robocodex};
            ControlManip~\cite{li2025controlmanip};
            OCI~\cite{wen2024object}, 
            ft_subsubsec]]
        [Enhanced Encoder for Manipulation (Sec. \ref{Sec:EEM}), ft_subsec
            [\textbf{Definition:} Strengthen the perception encoder to improve environmental understanding (e.g. 3D scene perception).,ft_subsubsec]
            [\textbf{Examples:} CEED-VLA~\cite{song2025ceed};
            ChatVLA-2~\cite{zhou2025vision};
            OE-VLA~\cite{zhao2025unveiling};
            NBAgent~\cite{liang2024never};
            RoboSpatial~\cite{song2024robospatial};
            ChatVLA~\cite{zhou2025chatvla};
            TinyVLA~\cite{wen2025tinyvla};
            MoLe-VLA~\cite{zhang2025mole};
            InSpire~\cite{zhang2025inspire};
            RoboBERT (Architecture)~\cite{wang2025robobert};
            Emma-X (Spatial Data Tuning)~\cite{sun2024emma},
            ft_subsubsec]]
        [Enhanced Representation for Manipulation (Sec. \ref{Sec:ERM}), ft_subsec
            [\textbf{Definition:} Refine task-relevant perceptual representations (i.e. introduce extra modalities) without changing the base encoder.,ft_subsubsec]
            [\textbf{Examples:} ForceVLA~\cite{yu2025forcevla};
            OmniVTLA~\cite{cheng2025omnivtla};
            VLA-Touch~\cite{bi2025vla};
            VTLA~\cite{zhang2025vtla};
            3D-CAVLA~\cite{bhat20253d};
            Evo-0~\cite{lin2025evo};
            SpatialVLA~\cite{qu2025spatialvla};
            GeoVLA~\cite{sun2025geovla};
            OG-VLA~\cite{singh2025og};
            RoboFlamingo-Plus~\cite{li2025roboflamingo};
            Spatial Traces~\cite{patratskiy2025spatial};
            CronusVLA~\cite{li2025cronusvla};
            MemoryVLA~\cite{shi2025memoryvla};
            RICL~\cite{sridhar2025ricl};
            TraceVLA~\cite{zheng2024tracevla};
            ObjectVLA~\cite{zhu2025objectvla};
            DexGraspVLA~\cite{zhong2025dexgraspvla};
            OTTER~\cite{huang2025otter};
            VLA-Cache~\cite{xu2025vla};
            BFA~\cite{lan2025bfa};
            VLAS~\cite{zhaovlas};
            LLARVA~\cite{niullarva};
            OpenVLA-OFT (Multi-camera Input)~\cite{kim2025fine};
            ControlManip (Object-centric Representation)~\cite{li2025controlmanip},
            ft_subsubsec]]
      ]
      [\textbf{Improving Embodiment\\Awareness (Sec. \ref{Sec:Improving Embodiment Awareness})}, ft_sec
        [Forward/inverse Kinematics\\Learning (Sec. \ref{Sec:Forward kinematics learning}), ft_subsec            
            [\textbf{Definition:}  Learn embodiment-specific mappings by predicting future proprioceptive states from actions or inferring actions for target states.,ft_subsubsec]
            [\textbf{Examples:} Zhang et al.~\cite{zhang2024effective};
            MIDAS~\cite{li2024mastering};
            DemoGen~\cite{jin2024learning}, ft_subsubsec]]
        [Action Head Designing (Sec. \ref{Sec. Action Head Designing}), ft_subsec
            [\textbf{Definition:}  Redesign the policy/output head to map multimodal features to low-level control commands.,ft_subsubsec]
            [\textbf{Examples:}TAIL~\cite{liu2023tail};
            OMLA~\cite{zhu2025efficient};
            DiffusionVLA~\cite{wen2025diffusionvla};
            Dita~\cite{hou2025dita};
            VOTE~\cite{lin2025vote};
            Pan et al.~\cite{pan2024vision};
            HybridVLA~\cite{liu2025hybridvla};
            Discrete Diffusion VLA~\cite{liang2025discrete};
            PD-VLA~\cite{song2025accelerating};
            $\pi_0 $-Fast~\cite{pertsch2025fast};
            OpenVLA-OFT (Parallel Decoding \& Action Chunking)~\cite{kim2025fine};
            VQ-VLA~\cite{wang2025vq};
            HPT~\cite{wang2024scaling};
            UniAct~\cite{zheng2025universal};
            CEED-VLA (Jacobi Decoding)~\cite{song2025ceed};
            CronusVLA (Action Adaptation)~\cite{li2025cronusvla},
            ft_subsubsec]]
      ]
      [\textbf{Deepening Task\\Comprehension (Sec. \ref{Sec:Deepening Task Comprehension})}, ft_sec
        [Human–Robot Interaction (Sec. \ref{Sec:HRI}), ft_subsec
            [\textbf{Definition:} Incorporate expert knowledge through online interactions (e.g. interventions or preference alignment) to improve safety and performance., ft_subsubsec]
            [\textbf{Examples:} SC-VLA~\cite{li2024self};
            Phoenix~\cite{xia2025phoenix};
            HAPO~\cite{xia2025robotic};
            ADC~\cite{huang2025adversarial};
            RoboCopilot~\cite{wu2025robocopilot};
            Expert-VLA~\cite{xiang2025vla};
            ConRFT~\cite{chen2025conrft};
            RAPL~\cite{tian2024maximizing};
            AMF (Expert Provides Demonstration)~\cite{bagatella2024active},
            ft_subsubsec]]
        [Hierarchical Task Manipulation (Sec. \ref{Sec:HTM}), ft_subsec
            [\textbf{Definition:} Decompose long-horizon tasks into sub-goals/primitives to enable structured planning and closed-loop execution., ft_subsubsec]
            [\textbf{Examples:} ECoT-Lite~\cite{chen2025training}; 
            ECoT~\cite{zawalskirobotic};
            Emma-X~\cite{sun2024emma};
            Fast ECoT~\cite{duan2025fast};
            Hi Robot~\cite{shihi};
            HiRT~\cite{zhang2025hirt};
            OneTwoVLA~\cite{lin2025onetwovla};
            ThinkAct~\cite{huang2025thinkact};
            STEER~\cite{smith2024steer};
            GRAPE~\cite{zhang2024grape};
            RoboMatrix~\cite{mao2024robomatrix};
            CLIP-RT~\cite{kang2024clip};
            RoboDexVLM~\cite{liu2025robodexvlm};
            PALO~\cite{myerspolicy};
            DiffusionVLA (Architecture \& CoT)~\cite{wen2025diffusionvla};
            DexVLA (Architecture \& Embodied Curriculum Learning)~\cite{wen2025dexvla};
            DexGraspVLA (Architecture)~\cite{zhong2025dexgraspvla};
            SC-VLA (Architecture)~\cite{li2024self};
            Phoenix (Architecture)~\cite{xia2025phoenix};
            HAMSTER (Architecture)~\cite{li2025hamster},
            ft_subsubsec]]
      ]
      [\textbf{Multi-component Integration (Sec. \ref{Sec:Mutiple Component Integration})}, ft_sec
      [Reinforcement Learning (Sec.~\ref{Sec. RL}), ft_subsec
            [\textbf{Definition:} Use reinforcement fine-tuning to align and improve the VLA models., ft_subsubsec]
            [\textbf{Examples:} Robot-R1~\cite{kim2025robot};
            SafeVLA~\cite{zhang2025safevla};
            TGRPO~\cite{chen2025tgrpo};
            ReinboT~\cite{zhang2025reinbot};
            RLDG~\cite{xu2024rldg};
            RFTF~\cite{shu2025rftf};
            VLA-RL~\cite{lu2025vla};
            LiFT~\cite{nam2023lift};
            IKER~\cite{patel2025real};
            OpenPAL~\cite{zhai2023building};
            Policy Decorator~\cite{yuan2024policy};
            RIPT-VLA~\cite{tan2025interactive};
            Liu et al.~\cite{liu2025can};
            LiReN~\cite{stachowiczlifelong};
            FLaRe~\cite{hu2024flare};
            iRe-VLA~\cite{guo2025improving};
            LaMo~\cite{shiunleashing};
            PA-RL~\cite{mark2024policy};
            V-GPS~\cite{nakamotosteering};
            KAGI (Online RL)~\cite{lee2024affordance};
            ConRFT (Online RL)~\cite{chen2025conrft};
            ThinkAct (GRPO Optimization)~\cite{huang2025thinkact};
            HAPO (Preference Optimization)~\cite{xia2025robotic};
            RAPL (Preference Optimization)~\cite{tian2024maximizing};
            GRAPE (Preference Optimization)~\cite{zhang2024grape},
            ft_subsubsec]]
      [Visual Interaction Prediction (Sec.~\ref{Sec. VIP}), ft_subsec
            [\textbf{Definition:} Predict future observations and interaction dynamics under candidate actions to guide planning and control., ft_subsubsec]
            [\textbf{Examples:} CoT-VLA~\cite{zhao2025cot};
            WorldVLA~\cite{cen2025worldvla};
            Moto~\cite{chen2024moto};
            DreamVLA~\cite{zhang2025dreamvla};
            FLARE~\cite{zheng2025flare};
            CoTDiffusion~\cite{ni2024generate};
            GENIMA~\cite{shridhar2025generative};
            HAMSTER~\cite{li2025hamster};
            Gr-2~\cite{cheang2024gr};
            VLMPC~\cite{chen2025vision};
            MineDreamer~\cite{zhou2024minedreamer};
            VPP~\cite{hu2024video};
            UP-VLA~\cite{zhang2025up};
            MPI~\cite{zeng2024learning};
            VIRT~\cite{li2024virt};
            DISCO~\cite{hao2024language};
            ReinboT (Image Decoder)~\cite{zhang2025reinbot}, 
            ft_subsubsec]]
      [Active Dataset Processing (Sec.~\ref{Sec. EFT}), ft_subsec
            [\textbf{Definition:} Actively curate and augment data to maximize information per sample and make post-training more sample-efficient., ft_subsubsec]
            [\textbf{Examples:} AMF~\cite{bagatella2024active};
            SWBT~\cite{wu2024swbt};
            DataPlatter~\cite{zheng2025dataplatter};
            Gao et al.~\cite{gao2024efficient};
            Lin et al.~\cite{lin2024data};
            DexVLA~\cite{wen2025dexvla};
            RoboBERT~\cite{wang2025robobert};
            SOAR~\cite{zhouautonomous}, ft_subsubsec]]]
    ]
  \end{forest}
  \caption{Taxonomy of post-training VLA models proposed in this study.}
  \label{fig:ft-taxonomy}
\end{figure*}
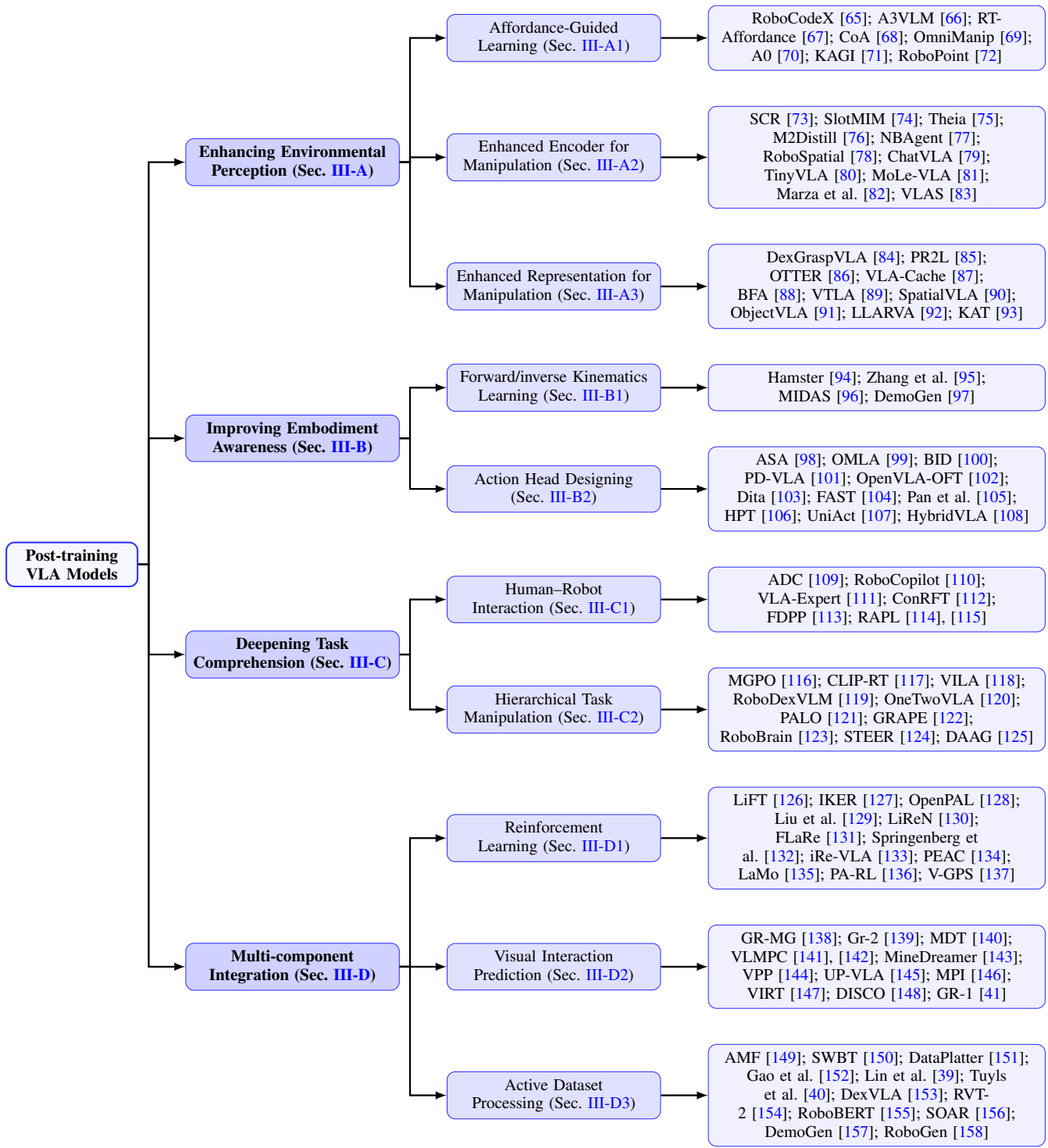

\section{Methodology for Review}

This review focuses on research within the scope of VLA models post-training. The process comprises two phases. In the first phase, a search was conducted on August 31, 2025, across IEEE Xplore, ACM Digital Library, Web of Science Core Collection, Scopus, ScienceDirect, and Google Scholar, including peer-reviewed conferences/journals and preprints written in English. The search string is as follows:
\begin{itemize}
    \item (``vision-language-action" OR "VLA") AND (``robotic manipulation" OR ``task execution") AND (``fine-tuning" OR ``model adaptation" OR ``post-training")
\end{itemize}

Given the rapidly evolving nature of the area, studies published after this date are excluded from the present review. The deliberately broad query maximizes coverage and yields a total of 997 records.

In the second phase, records (titles, abstracts, and full texts) are independently screened by three reviewers (R1-R3) to assess relevance to VLA post-training for robotic manipulation. Studies not centered on VLA models for manipulation are excluded. Papers whose primary focus is pre-training, architecture design, benchmarking, or secondary reviews are also excluded. Items with unanimous inclusion (95, 3/3) are admitted directly; items with majority inclusion (71, 2/3) are discussed to reach consensus. This process yields 129 research studies included in this review.

The detailed results are as follows: R1 keeps 160 records, R2 keeps 307, while R3 keeps 133 records. Inter-rater reliability indicated acceptable agreement (Cohen's Kappa: 0.48 for R1-R2, 0.62 for R1-R3, 0.44 for R2-R3; Fleiss's Kappa: 0.49; Percent agreement: 0.81 for R1-R2, 0.91 for R1-R3, 0.80 for R2-R3). These results support the reliability of the screening procedure. The full list of included papers and screening decisions is provided in the supplementary material.

\section{Taxonomy for Post-training VLA Models}

This section presents a comprehensive taxonomy for post-training VLA models from the perspective of human motor learning (Fig.~\ref{fig:ft-taxonomy}), organized around three key components: environments, embodiments, and tasks.

\subsection{Enhancing Environmental Perception}

\begin{figure*}[tb]
   \centering
     \includegraphics[width=7in]{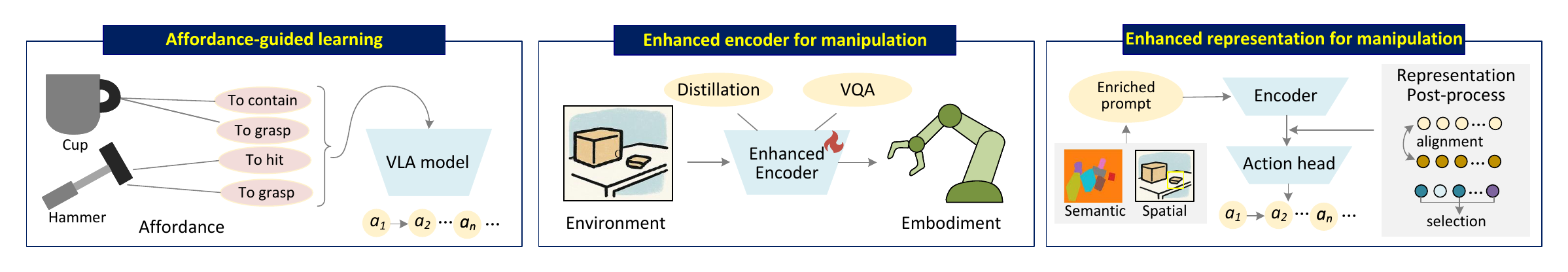}
       \caption{Summary of techniques for enhancing environmental perception: affordance-guided learning (Sec.~\ref{Sec:Affordance-Guided Learning}); enhanced encoder for manipulation (Sec.~\ref{Sec:EEM}); enhanced representation for manipulation (Sec.~\ref{Sec:ERM}).}
    \centering
    \label{fig:EEP}
\end{figure*}

\label{Sec:Enhancing Environmental Perception}

\begin{table*}[!htb]\scriptsize
\centering
\renewcommand{\arraystretch}{1.2}
\caption{Representative techniques for enhancing environmental perception.}
\begin{threeparttable}
\begin{tabular}{
  >{\centering\arraybackslash}m{1.8cm}
  >{\centering\arraybackslash}m{1cm}
  >{\arraybackslash}m{12cm}
}
\toprule
\textbf{Methods} & \textbf{Type} & \textbf{Key ideas} \\
\midrule
COA~\cite{li2024improving}
& AGL
& Links multiple object affordances into a chain that supplies intermediate sub-goals, guiding the policy. \\
\cline{3-3}
A3VLM~\cite{huanga3vlm}
& AGL
& Incorporates object-centric affordance representations into the VLM decoder to generate robot actions. \\
\cline{3-3}
RoboPoint~\cite{yuanrobopoint}
& AGL
& Post-trains a VLM to extract spatial affordances of target objects and guide manipulation. \\
\midrule
CEED-VLA~\cite{song2025ceed}
& EEM
& Integrates consistency distillation with an early-exit strategy to improve decoding efficiency. \\
\cline{3-3}
RoboSpatial~\cite{song2024robospatial}
& EEM
& Utilizes visual question answering to enhance the visual encoder’s 3D perception and cognition.\\
\cline{3-3}
MoLe-VLA~\cite{zhang2025mole}
& EEM
& Employs dynamic routing to skip selected encoder layers, reducing computation while retaining perception quality. \\
\midrule
Otter~\cite{huang2025otter}
& ERM
& Aligns task-aware visual embeddings with task instruction to strengthen semantic understanding. \\
\cline{3-3}
VTLA~\cite{zhang2025vtla}
& ERM
& Augments the VLA with tactile sensing, improving perception in contact-rich interactions. \\
\cline{3-3}
SpatialVLA~\cite{qu2025spatialvla}
& ERM
& Injects 3D information into input observations to enhance spatial perception. \\
\bottomrule
\end{tabular}
\label{table:Environment}
\begin{tablenotes}
    \item[1] Abbreviations: AGL: affordance-guided learning; EEM: enhanced encoder for manipulation; ERM: enhanced representation for manipulation.
\end{tablenotes}
\end{threeparttable}
\end{table*}

\subsubsection{Human Strategies for Enhancing Environmental Perception}

\label{Sec: human enhance environmental perception}

During motor learning, humans employ multiple strategies to improve environmental perception. One strategy leverages environmental affordances. In Gibson’s ecological psychology, an affordance denotes the action possibilities provided by objects and layouts~\cite{gibson1966senses,bach2014affordance}. Humans can naturally detect such cues, allowing the environment to guide feasible actions; these cues accelerate skill acquisition and support the effective use of prior knowledge~\cite{bach2014affordance}.

Humans also refine internal perceptual encodings through experience. With practice, the sensorimotor system updates internal models to maintain calibration with environmental changes~\cite{avraham2022contextual}. Prior experiences shape context-specific sensory representations that mitigate skill forgetting and speed up learning~\cite{forano2021direct,ferrari2022updating}.

A complementary strategy is multi-sensory integration. Humans combine information from multiple modalities to obtain more reliable estimates of the environment, yielding redundancy and increased robustness~\cite{camponogara2021integration}. Evidence from grasping shows that adding a modality alongside vision can maintain accuracy and confidence at comparable levels~\cite{camponogara2021integration}. Supplementary guidance and explicit instruction can further facilitate understanding of environmental structure and speed the learning process~\cite{taylor2014explicit}.

Drawing on insights from human motor learning, following VLA post-training directions are considered: (i) affordance-guided learning (Sec.~\ref{Sec:Affordance-Guided Learning}), (ii) enhanced encoders for manipulation (Sec.~\ref{Sec:EEM}), and (iii) enhanced representations for manipulation (Sec.~\ref{Sec:ERM}).

\subsubsection{Affordance-Guided Learning}

\label{Sec:Affordance-Guided Learning}

Affordances, introduced by Gibson in a psychology study in mid–20th century, describe the actionable possibilities offered by the environment~\cite{gibson1966senses}. Before the emergence of VLA models, affordance has already been extensively explored in robotic manipulation~\cite{yamanobe2017brief}. In this context, affordance denotes the interaction between agent capabilities and environmental features that gives rise to behavior~\cite{stoffregen2018affordances}. Because affordance captures object–environment relationships critical for manipulation, affordance-guided methods are presented in Sec.~\ref{Sec:Enhancing Environmental Perception}.

Affordances for robotic manipulation can be grouped into two categories: (i) manipulation-related, which encode strategies for robot-object interaction (the \emph{how})~\cite{huanga3vlm, li2025controlvla}, and (ii) spatial, which convey positional cues indicating where manipulation should occur (the \emph{where})~\cite{li2024improving}. Recent studies embed affordance knowledge within VLA models to improve manipulation performance~\cite{li2024improving, huanga3vlm, yuanrobopoint, pan2025omnimanip, nasirianyrt, xu2025a0, wen2024object, li2025controlmanip}.

Some approaches utilize only spatial affordances to produce guidance signals, whereas many leverage both manipulation-related and spatial cues inherent to objects~\cite{huanga3vlm, yuanrobopoint, pan2025omnimanip, nasirianyrt, xu2025a0}. A common implementation encodes affordance information directly within language instructions, enabling explicit affordance transfer via enriched prompts~\cite{pan2025omnimanip, yuanrobopoint, huanga3vlm, wen2024object}. Alternatively, pre-trained VLM can be fine-tuned on dedicated affordance datasets, supplying affordance representations for action outputs~\cite{xu2025a0, nasirianyrt, mu2024robocodex, li2025controlmanip}. Affordance knowledge can also be embedded within chain-of-thought (CoT) procedures to structure reasoning from spatial perception to manipulation~\cite{li2024improving}. Representative methods regarding affordance-guided learning are summarized in Table~\ref{table:Environment}.

\subsubsection{Enhanced Encoder for Manipulation}

\label{Sec:EEM}

Optimizing the perception encoder to refine environmental perception is a direct avenue for improving manipulation performance in VLA models. Even simple supervised fine-tuning with a fully unfrozen encoder can contribute to the manipulation success rates~\cite{zhang2024effective}. This subsection reviews strategies that strengthen the encoder’s perceptual capacity.

Knowledge distillation enhances perception in VLA models by transferring representations from a high-capacity teacher to a student, mitigating catastrophic forgetting during supervised fine-tuning~\cite{liang2024never, song2025ceed}. Distillation can also convey specialized skills, such as 3D visual understanding, from strong teachers to the base encoder~\cite{liang2024never}. Efficiency gains are achievable when a compact student is trained with distillation objectives, enabling faster decoding with comparable accuracy~\cite{song2025ceed}.

Because VLA models are inherently multimodal, visual question answering (VQA) offers a natural supervision signal for improving encoder~\cite{song2024robospatial, zhou2025chatvla, zhou2025vision}. VQA-based training strengthens specific perceptual capabilities such as 3D understanding~\cite{song2024robospatial} and improves overall multimodal environmental awareness~\cite{zhou2025chatvla, zhou2025vision}. It should be noted that VQA is complementary to other training strategies and can be combined with distillation or architectural refinements~\cite{zhou2025chatvla, zhou2025vision}.

Architectural refinements can also improve the encoder. Representative directions include lightweight or dynamic designs~\cite{wen2025tinyvla, zhang2025mole} and the incorporation of additional information~\cite{zhao2025unveiling}. These adjustments either reduce computational cost or expand the environmental understanding of the perception module. Representative methods are summarized in Table~\ref{table:Environment}.

\subsubsection{Enhanced Representation for Manipulation}

\label{Sec:ERM}

Enhancing task-relevant environmental representations supports the policy head in action inference. Unlike the methods in Sec.~\ref{Sec:EEM}, the approaches considered here primarily employ external modules or additional modalities, such as enriched prompts~\cite{zhu2025objectvla, niullarva}, representation selection or alignment~\cite{huang2025otter, lan2025bfa}, and tactile or depth signals~\cite{zhang2025vtla, li2025roboflamingo}, rather than relying solely on vision-language encoder fine-tuning.

A direct avenue is to incorporate modalities that are informative for manipulation but absent from standard VLA configurations. Examples include tactile sensing~\cite{yu2025forcevla} and depth cues~\cite{patratskiy2025spatial}. Tactile signals benefit contact-rich tasks~\cite{yu2025forcevla, cheng2025omnivtla, bi2025vla, zhang2025vtla}, whereas depth information strengthens spatial reasoning~\cite{bhat20253d, lin2025evo, qu2025spatialvla, sun2025geovla, singh2025og, li2025roboflamingo, patratskiy2025spatial, zhang2025inspire}. Incorporating these capabilities into VLA models has been shown to improve manipulation performance. Additional interaction channels can also aid usability; for instance, speech can improve robot control experience even if its direct impact on physical manipulation may be limited~\cite{zhaovlas}.

Another strategy uses more informative prompts to guide the vision-language encoder toward task-relevant representations~\cite{niullarva, zhu2025objectvla, qu2025spatialvla, zhong2025dexgraspvla, zheng2024tracevla}, analogous to prompt engineering in large language models~\cite{sahoo2024systematic}. Many studies leverage external models (e.g., SAM~\cite{kirillov2023segment}) to extract environmental features that are then incorporated into prompts fed to the VLA. Notably, prompts need not be purely natural language; non-linguistic or structured forms are also effective~\cite{zheng2024tracevla}.

A complementary line of work post-processes extracted vision-language representations~\cite{huang2025otter, lan2025bfa, xu2025vla}. For example, Huang et al.~\cite{huang2025otter} propose an alignment module that synchronizes visual embedding with instruction-derived representations to emphasize task-relevant features. Lan et al.~\cite{lan2025bfa} propose a fusion module that scores representations from multiple inputs and aggregates them according to these scores. Historical information can also enhance manipulation representations~\cite{li2025cronusvla, shi2025memoryvla, sridhar2025ricl}. Both long-term memory buffers~\cite{shi2025memoryvla} and short-term historical sequences~\cite{li2025cronusvla, sridhar2025ricl} provide context that improves decision quality. Representative methods in this section are summarized in Table~\ref{table:Environment}.

\subsection{Improving Embodiment Awareness}

\begin{figure}[tb]
   \centering
     \includegraphics[width=3.5in]{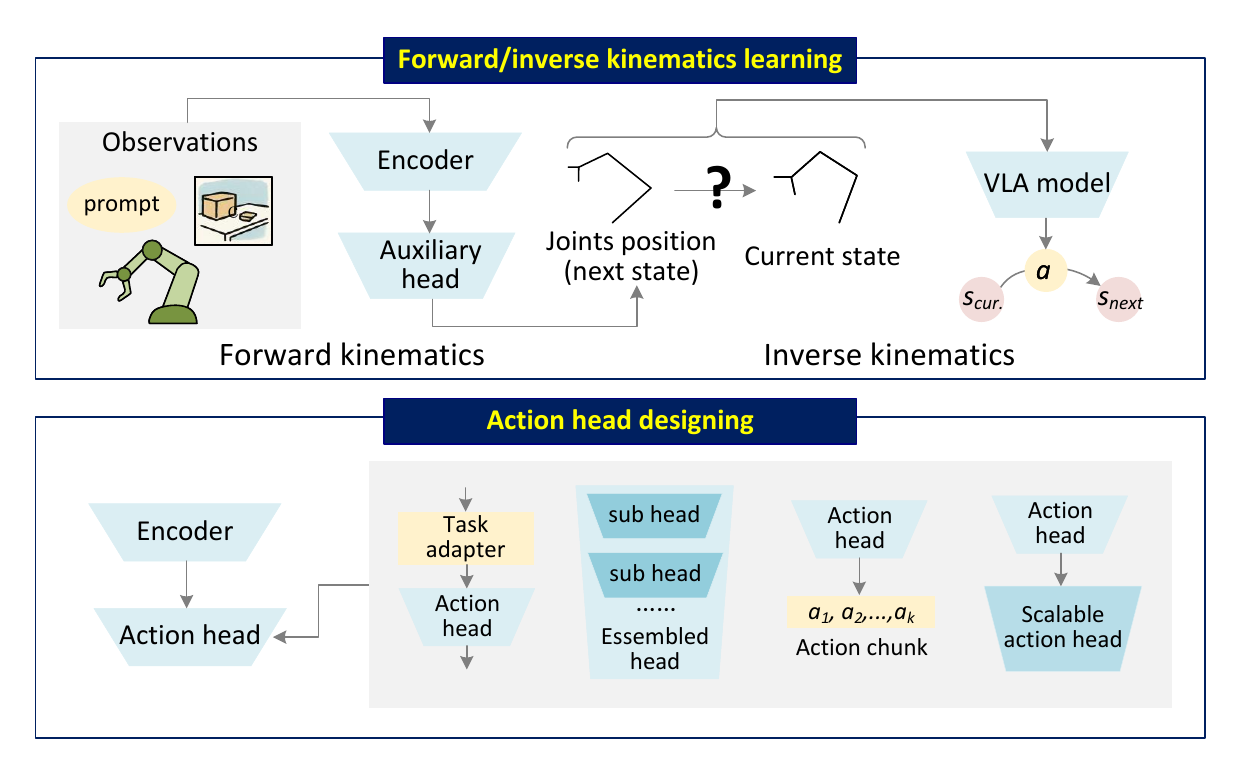}
       \caption{Summary of techniques for improving embodiment awareness: forward/inverse kinematics learning (Sec.~\ref{Sec:Forward kinematics learning}); action head designing (Sec.~\ref{Sec. Action Head Designing}).}
    \centering
    \label{fig:IEA}
\end{figure}

\label{Sec:Improving Embodiment Awareness}

\begin{table*}[!htb]\scriptsize
\centering
\renewcommand{\arraystretch}{1.2}
\caption{Representative techniques for enhancing embodiment awareness.}
\begin{threeparttable}
\begin{tabular}{
  >{\centering\arraybackslash}m{2cm}
  >{\centering\arraybackslash}m{1cm}
  >{\arraybackslash}m{13cm}
}
\toprule
\textbf{Methods} & \textbf{Type} & \textbf{Key ideas} \\
\midrule
Zhang et al.~\cite{zhang2024effective} 
& FK 
& Adds an $L_1$ auxiliary objective to predict end-effector properties (e.g., position, state). \\
\cline{3-3}
LLARVA~\cite{niullarva} 
& FK 
& Forecasts 2D end-effector trajectories from visual input within the VLA. \\
\cline{3-3}
Li et al.~\cite{li2024mastering} 
& IK 
& Maps image observations to actions for inverse-kinematics learning. \\
\midrule
Dita~\cite{hou2025dita} 
& AHD
& Employs a transformer-based diffusion head to denoise action trajectories, enabling scalable generation. \\
\cline{3-3}
DiffusionVLA~\cite{wen2025diffusionvla}
& AHD
& Adds a FiLM module that fuses vision-language reasoning into the action head. \\
\cline{3-3}
HybridVLA~\cite{liu2025hybridvla}
& AHD
& Combines autoregressive and diffusion action heads within a VLA to exploit complementary strengths. \\
\cline{3-3}
Fast~\cite{pertsch2025fast}
& AHD
& Proposes compression-based tokenization scheme for robot actions generation based on the discrete cosine transform. \\
\bottomrule
      \end{tabular}

\label{table:Embodiment}
\begin{tablenotes}
    \item[1] Abbreviations: Forward kinematics: FK; Inverse kinematics: IK; Action head designing: AHD.
\end{tablenotes}
\end{threeparttable}
\end{table*}

\subsubsection{Human Strategies for Enhancing Embodiment Awareness}

\label{Sec: human enhance embodiment awareness}

Embodiment awareness in humans is grounded in internal models, including forward and inverse models, within the motor system~\cite{de2008neural,imamizu2000human}. A forward model predicts the sensory consequences and future limb state given a motor command, enabling anticipatory control and movement stability~\cite{blakemore2000can}. An inverse model computes the motor commands required to attain a desired state. Together, these models establish an internal action perception loop that supports body awareness and sensorimotor integration, allowing fluid movement while attributing predicted sensations to the self.

Beyond this basic scheme, specialized neural controllers further refine embodiment awareness. The cerebellum is hypothesized to host multiple internal model modules tuned to different contexts or tools~\cite{imamizu2007reorganization}; during tool learning, new internal models are acquired, indicating that execution can recruit multiple models. The basal ganglia contribute to assembling discrete motor primitives into fluent action sequences via ``chunking''~\cite{graybiel1998basal}, supporting efficient selection and execution. Collectively, these controllers enrich the body schema and promote a sense of ownership and agency over skilled actions.

These insights motivate two directions for VLA post-training aimed at enhancing embodiment awareness: (i) forward/inverse kinematics learning (Sec.~\ref{Sec:Forward kinematics learning}), and (ii) action-head design (Sec.~\ref{Sec. Action Head Designing}).

\subsubsection{Forward/Inverse Kinematics Learning}

\label{Sec:Forward kinematics learning}

Kinematics in robotic systems comprises forward and inverse processes. Forward kinematics predicts the resulting pose from a given action, whereas inverse kinematics computes the action required to achieve the target pose. Because kinematic relations vary across embodiments, it is essential to enable a VLA model to internalize embodiment-specific mappings.

In forward kinematics learning, the VLA model forecasts future proprioceptive information, such as joint angles and end-effector positions and trajectories, from current observations and task instructions~\cite{zhang2024effective, li2024mastering, niullarva}. Predicting these signals enables implicit acquisition of the embodiment’s forward mapping. This is typically implemented via an auxiliary prediction head that maps fused representations to proprioceptive outputs. In inverse kinematics learning, the model infers actions that induce observed state transitions~\cite{jin2024learning}. By predicting embodiment-specific actions from sensory inputs and instructions, the model learns the inverse action generation mapping of the given embodiment. Representative methods are summarized in Table~\ref{table:Embodiment}.

\subsubsection{Action Head Designing}

\label{Sec. Action Head Designing}

The action head in VLA models maps vision-language representations to manipulation commands, translating task understanding and environmental perception into motor actions. Because embodiment-specific action spaces vary across platforms, action-head design is critical for embodiment awareness.

One family of designs employs adapters that reparameterize the multimodal encoder for action decoding~\cite{zhu2025efficient, liu2023tail}. Zhu et al.~\cite{zhu2025efficient} introduce lightweight adapters that support rapid task adaptation across diverse manipulation scenarios. Liu et al.~\cite{liu2023tail} study parameter efficient fine-tuning with task-specific adapters for pre-trained models, achieving fast adaptation while mitigating catastrophic forgetting.

Architectural design provides another line of improvement. Hou et al.~\cite{hou2025dita} report scaling behavior in the policy (action) head of VLA models, while Wen et al.~\cite{wen2025diffusionvla} use a FiLM conditioning module to fuse vision-language reasoning with action generation. Beyond a single policy head, multiple policy outputs can be fused into an ensemble, either temporally~\cite{lin2025vote} or across heterogeneous heads~\cite{liu2025hybridvla, pan2024vision}.

Efficiency and generalization are additional focuses. Action decoding efficiency can be improved via parallel decoding mechanisms~\cite{song2025accelerating, kim2025fine} and designed tokenization schemes~\cite{pertsch2025fast, wang2025vq}, and these routes can be combined~\cite{liang2025discrete}. For cross-embodiment generalization, some studies share a common latent representation while assigning embodiment-specific heads to different robot platforms, enabling distinct decoding pathways from a unified representation~\cite{wang2024scaling, zheng2025universal}. Representative action head designs are summarized in Table~\ref{table:Embodiment}.

\subsection{Deepening Task Comprehension}

\begin{figure}[tb]
   \centering
     \includegraphics[width=3.5in]{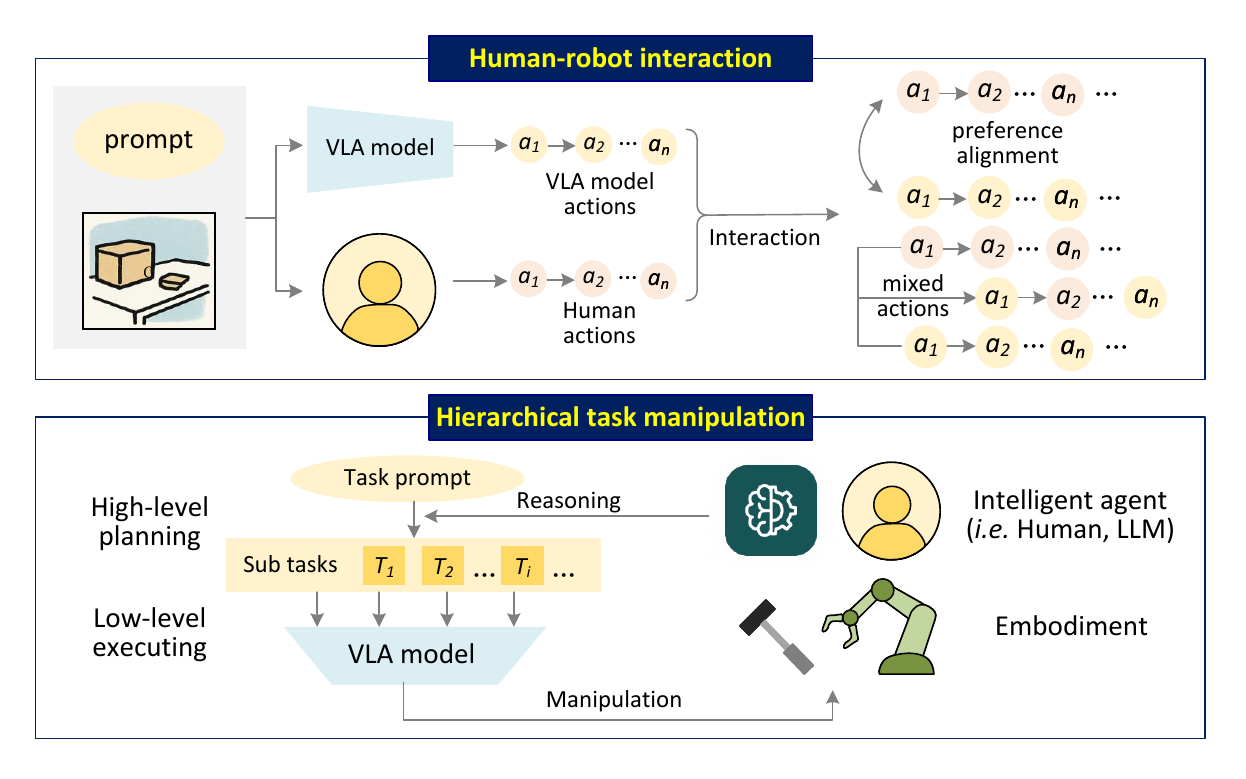}
       \caption{Summary of techniques for deepening task comprehension: human–robot interaction (Sec.~\ref{Sec:HRI}); hierarchical task manipulation (Sec.~\ref{Sec:HTM}).}
    \centering
    \label{fig:DTC}
\end{figure}

\label{Sec:Deepening Task Comprehension}

\begin{table*}[!htb]\scriptsize
\centering
\renewcommand{\arraystretch}{1.2}
\caption{Representative techniques for deepening task comprehension.}
\begin{threeparttable}
\begin{tabular}{
  >{\centering\arraybackslash}m{2cm}
  >{\centering\arraybackslash}m{1cm}
  >{\arraybackslash}m{14cm}
}
\toprule
\textbf{Methods} & \textbf{Type} & \textbf{Key ideas} \\
\midrule
VLA-Expert~\cite{xiang2025vla}
& HRI
& Mixes sparse expert actions with policy outputs during execution, then fine-tunes on interactions to improve the VLA model. \\
\cline{3-3}
ConRFT~\cite{chen2025conrft}
& HRI
& Combines expert oversight with online RL to enforce safety and correctness, and leverages collected interactions for post-training. \\
\cline{3-3}
ADC~\cite{huang2025adversarial}
& HRI
& Uses adversarial task/object variants to elicit informative interventions and high–prior-density demonstrations. \\
\cline{3-3}
RAPL~\cite{tian2024maximizing}
& HRI
& Aligns internal policy with human visual feedback during manipulation to encode expert preferences. \\
\midrule
CLIP-RT~\cite{kang2024clip}
& HTM
& Training sub-task instructions and action primitives in CLIP-style~\cite{radford2021learning}, then aligns visual observations with instructions during inference to select appropriate primitives for closed-loop control. \\
\cline{3-3}
GRAPE~\cite{zhang2024grape}
& HTM
& Decomposes trajectories into discrete temporal stages to enable multi-scale alignment. \\
\cline{3-3}
ECoT~\cite{zawalskirobotic}
& HTM
& Integrates CoT prompting into the robotic stack to realize embodied CoT and improve long-horizon manipulation. \\
\bottomrule
\end{tabular}
\label{table:Task}
\begin{tablenotes}
    \item[1] Abbreviations: Human-robot interaction: HRI; Hierarchical task manipulation: HTM.
\end{tablenotes}
\end{threeparttable}
\end{table*}

\subsubsection{Human Strategies for Deepening Task Comprehension}

\label{Sec: human Deepening Task Comprehension}

Human strategies for deepening task comprehension center on observational learning from abstracted rules and hierarchical organization of action. Humans acquire complex skills through instructional guidance and expert demonstrations, reducing reliance on trial-and-error and improving safety and sample efficiency~\cite{csibra2011natural}. A related neural mechanism shows activation of the mirror neuron system during action observation, enabling internal simulation and facilitating understanding of demonstrated behaviors~\cite{cattaneo2009mirror}. Beyond imitation, advanced cognitive and perceptual capacities support the formation of high-level task representations and abstract rules~\cite{merel2019hierarchical}. Humans can infer underlying principles and goals, enabling quick adaptation of acquired skills to novel contexts~\cite{friedman2019observation}.

Another strategy in human motor learning is the use of hierarchical organization to manage long-horizon complex tasks. When faced with these challenging tasks, humans naturally decompose the task into simpler sub-tasks. Research in neuroscience confirms that chunking a sequence of actions is an efficient way to solve long action sequences: breaking a complex movement into a series of shorter segments reduces the computational complexity of planning and control~\cite{ramkumar2016chunking}. This hierarchical control structure also mirrors the organization of the human nervous system, where high-level circuits plan actions and low-level circuits execute motor commands~\cite{merel2019hierarchical}.

Drawing on these findings, VLA post\-/training strategies for deepening task comprehension are organized into: (i) human-robot interaction (Sec.~\ref{Sec:HRI}) and (ii) hierarchical task manipulation (Sec.~\ref{Sec:HTM}).

\subsubsection{Human-Robot Interaction}
\label{Sec:HRI}

Current manipulation tasks for VLA models remain relatively naive from the perspective of human experts. Transferring expert task knowledge via interactions is therefore a promising direction. Such transfer can be realized through explicit or implicit processes.

Explicit methods place human experts in the control loop, using interaction data to post-train VLA policies and improve efficiency, success rates, and safety~\cite{chen2025conrft, xiang2025vla, wu2025robocopilot, li2024self, xia2025phoenix}. A defining feature is direct collaboration between experts and the VLA policy; a common protocol triggers human takeover after failure, logs corrective demonstrations, and then updates the policy to recover previously unsuccessful tasks~\cite{wu2025robocopilot, li2024self, xia2025phoenix}. Expert behavior can also be mixed sparsely during execution; VLA-Expert~\cite{xiang2025vla} injects a small fraction of expert actions into a pre-trained VLA to boost performance. Human supervision further enhances safety in online learning; ConRFT~\cite{chen2025conrft} combines expert guidance with online reinforcement learning to enforce safe and correct actions.

Implicit methods encode expertise by aligning to human preferences or by embedding expert knowledge within demonstrations~\cite{rafailov2023direct, tian2024maximizing, huang2025adversarial}. Preference alignment (e.g., direct preference optimization) adjusts the policy to match expert choices~\cite{rafailov2023direct}. Representation alignment maps human attention patterns during manipulation to VLA internal representations~\cite{tian2024maximizing}. Demonstration-based alignment, as in Huang et al.~\cite{huang2025adversarial}, leverages expert interventions to generate corner-case trajectories with high prior density, improving sample efficiency. Explicit and implicit alignment can be combined: data collected in online interaction can subsequently be used offline within preference alignment frameworks~\cite{xia2025robotic}. Representative techniques are summarized in Table~\ref{table:Task}.

\subsubsection{Hierarchical Task Manipulation}

\label{Sec:HTM}

Hierarchical task decomposition draws on human problem-solving strategies, whereby complex, long-horizon objectives are partitioned into manageable sub-goals~\cite{kahneman2011thinking, voytek2015oscillatory}. Effective decomposition requires a model with a strong understanding of the manipulation tasks; accordingly, hierarchical structures for manipulation are treated here as methods for deepening task comprehension.

An intuitive realization introduces a high-level planner to segment long-horizon tasks and a low-level controller to execute sub-goals~\cite{shihi, zhang2025hirt, lin2025onetwovla, huang2025thinkact, li2025hamster}. The planner can be foundation models~\cite{zhang2024grape, ji2025robobrain} or human experts~\cite{zheng2025dataplatter, smith2024steer}. Decomposition is diverse: beyond the stepwise style, tasks with strong spatial-reasoning demands may be partitioned into a spatial-analysis phase and a physical-interaction phase to enable targeted data collection~\cite{zheng2025dataplatter}.

Another line introduces CoT prompting to enable embodied CoT reasoning~\cite{chen2025training, zawalskirobotic, sun2024emma, duan2025fast, zhao2025cot, ni2024generate}. Although CoT is not itself a hierarchical controller, coupling CoT with the robot control stack yields an effective high- and low-level structure. CoT has improved VLA performance on long-horizon manipulation~\cite{zawalskirobotic, sun2024emma}, while efficient integration with robotic systems remain open problems~\cite{chen2025training, duan2025fast}.

A complementary direction emphasizes control primitives and fine-grained sub-task specifications: extracting reusable manipulation primitives~\cite{mao2024robomatrix, kang2024clip, liu2025robodexvlm} or describing sub-tasks at higher resolution~\cite{myerspolicy}. During execution, primitive invocation supports online adaptation and closed-loop control, improving sub-goal sequencing~\cite{kang2024clip, liu2025robodexvlm, mao2024robomatrix}. Representative hierarchical decomposition methods are summarized in Table~\ref{table:Task}.

\subsection{Multi-component Integration}

\label{Sec:Mutiple Component Integration}

\begin{figure*}[tb]
   \centering
     \includegraphics[width=7in]{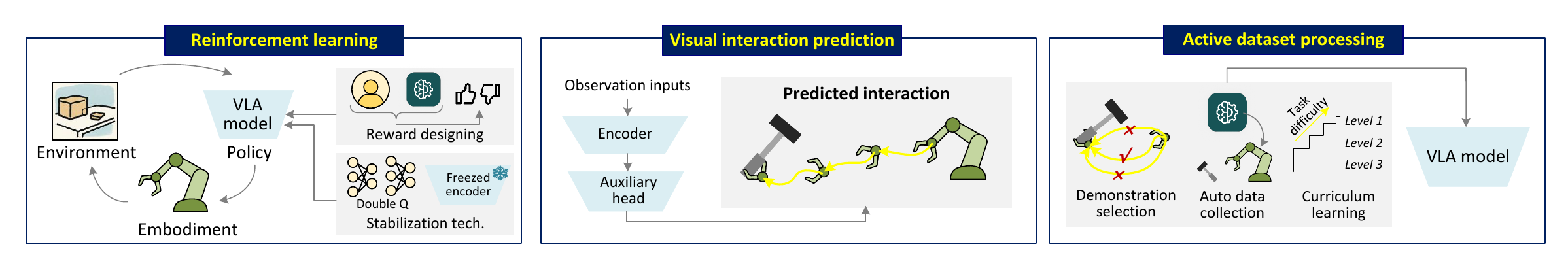}
       \caption{Summary of techniques for multi-component integration: reinforcement learning (Sec.~\ref{Sec. RL}); visual interaction prediction (Sec.~\ref{Sec. VIP}); active dataset processing (Sec.~\ref{Sec. EFT}).}
    \centering
    \label{fig:MCI}
\end{figure*}

\begin{table*}[!htb]\scriptsize
\centering
\renewcommand{\arraystretch}{1.2}
\caption{Representative techniques for involving multi-component of the VLA model learning.}
\begin{threeparttable}
\begin{tabular}{
  >{\centering\arraybackslash}m{1.6cm}
  >{\centering\arraybackslash}m{0.8cm}
  >{\arraybackslash}m{14.5cm}
}
\toprule
\textbf{Methods} & \textbf{Type} & \textbf{Key ideas} \\
\midrule
PA-RL~\cite{mark2024policy}
& RL
& Trains an offline RL critic to score actions sampled from a base policy, re-ranks selections by Q-value, and distills them to refine the policy. \\
\cline{3-3}
IKER~\cite{patel2025real}
& RL
& Leverages VLM to generate and refine reward functions, providing dense task feedback for RL. \\
\cline{3-3}
ConRFT~\cite{chen2025conrft}
& RL
& Integrates human-in-the-loop interaction and behavioral cloning within a two-stage offline-online RL. \\
\midrule
VLMPC\ \ \ \ \ \ \ \ ~\cite{zhao2024vlmpc,chen2025vision}
& VIP
& Uses an action-conditioned video predictor within an MPC objective, minimizing pixel-wise distance between forecast frames and a goal image to optimize control. \\
\cline{3-3}
VPP~\cite{hu2024video}
& VIP
& Adapts a pre-trained video generator on manipulation data, then trains an inverse dynamics model conditioned on the prediction to infer actions. \\
\cline{3-3}
MPI~\cite{zeng2024learning}
& VIP
& Predicts interaction frames between start and goal states and detects objects/interaction regions in the forecasts to guide action selection. \\
\midrule
AMF~\cite{bagatella2024active}
& ADP
& Selects demonstrations for active post-training by information gain relative to the target policy. \\
\cline{3-3}
SSDF~\cite{wu2024swbt}
& ADP
& Augments the fine-tuning set with imperfect (failed) manipulation trials. \\
\cline{3-3}
Gao et al.~\cite{gao2024efficient}
& ADP
& Factorizes tasks by object/scene properties and uses decomposition-based collection to improve zero-shot generalization to unseen combinations. \\
\cline{3-3}
SOAR~\cite{zhouautonomous}
& ADP
& Uses foundation-model for autonomous data collection by the target policy, enabling self-supervised refinement from gathered demonstrations. \\
\bottomrule

\end{tabular}

\label{table:Muti}
\begin{tablenotes}
    \item[1] Abbreviations: Reinforcement learning: RL; Visual interaction prediction: VIP; Active Dataset Processing: ADP.
\end{tablenotes}
\end{threeparttable}

\end{table*}

\subsubsection{Human Strategies for Comprehensive Improvement}

\label{Sec: human comprehensive improvement}

Beyond isolated component gains, human motor learning benefits in a comprehensive way. For example, skills can be refined through trial-and-error shaped by both reward-based and error-based feedback~\cite{kasdin2025natural,truong2023error}. Circuits in the basal ganglia encode reward prediction errors that reinforce successful actions, whereas the cerebellum and related sensorimotor networks compute sensory prediction errors to update motor commands. This joint adaptation provides a comprehensive improvement during learning.

In parallel, comprehensive internal models support the learning. Beyond forward and inverse models, neurophysiological evidence indicates that the cerebellum integrates the copy of motor commands with sensory feedback, while higher-level cortical systems support inverse-model learning, action planning, and sequencing. These phenomena are often described as ``internal world models'' maintained by the brain, which simulate action outcomes and enable rapid adaptation when task demands or environmental contexts change~\cite{diester2024internal}.

Human skill acquisition also benefits from structured, variable practice. Rather than repeating an identical movement, exposure to diverse, informative experiences produces more robust learning. A representative example is the contextual interference effect: practicing a motor skill under a high-interference schedule (randomly interleaving variations) can slow initial acquisition but yields superior retention and transfer~\cite{czyz2024high}. Thus, structured schedules strengthen motor memory formation and improve learning efficiency.

Motivated by these observations, comprehensive VLA post-training is organized around: (i) reinforcement learning (Sec.~\ref{Sec. RL}), (ii) visual interaction prediction (Sec.~\ref{Sec. VIP}), and (iii) active dataset processing (Sec.~\ref{Sec. EFT}).

\subsubsection{Reinforcement Learning}

\label{Sec. RL}

RL has been widely adopted in robotic research~\cite{kober2013reinforcement}. More recently, RL frameworks—particularly reinforcement learning from human feedback (RLHF)—have demonstrated efficacy in foundation model~\cite{kaufmann2023survey}. These advances have inspired the application of RL in VLA models. In RL, an agent iteratively interacts with environments under task-specific rewards, enabling policy optimization. Accordingly, RL naturally couples the three fundamental components of VLA learning: environments, embodiments, and tasks.

Reinforcement fine-tuning can learn policies from suboptimal demonstrations while maintaining generalization~\cite{liu2025can, mark2024policy}, but instability during RL-based post-training remains a key challenge~\cite{hu2024flare, stachowiczlifelong, guo2025improving, chen2025conrft, nakamotosteering, mark2024policy, yuan2024policy}. Two primary causes are (i) the transition from supervised pre-training to RL post-training and (ii) Q-function overestimation~\cite{hu2024flare, stachowiczlifelong}. Mitigations include behavior-cloning regularization (via an auxiliary BC loss~\cite{hu2024flare, yuan2024policy, chen2025conrft} or alternating supervised and RL updates~\cite{guo2025improving}) and freezing the vision-language encoder to stabilize features~\cite{chen2025conrft, guo2025improving}). Established RL stabilizers such as clipped double-Q learning~\cite{fujimoto2018addressing}, ensemble critics~\cite{chenrandomized}, and critic normalization~\cite{ball2023efficient} are also effective; Stachowicz et al.~\cite{stachowiczlifelong} integrate these techniques in a lifelong-learning framework. Additionally, critics can be trained offline and used online for action re-ranking~\cite{mark2024policy, nakamotosteering}, with newly collected data further refining both policy and critic~\cite{mark2024policy}.

Sparse rewards in manipulation poses another challenge in RL-based post-training. A straightforward approach is to use foundation models to produce dense reward signals, reducing handcrafted reward engineering~\cite{nam2023lift, patel2025real, zhai2023building, lu2025vla, lee2024affordance}. Rewards may be inferred directly from language instructions and visual observations~\cite{nam2023lift, lee2024affordance, zhai2023building, lu2025vla} or specified via programmatic functions~\cite{patel2025real}. Existing techniques designed for sparse-reward RL can also improve efficacy, such as dynamic rollout sampling, leave-one-out advantage estimation, and temporal credit assignment to densify feedback~\cite{lu2025vla, tan2025interactive}.
 
RL-based training also supports alignment to specific preferences: user preferences~\cite{chen2025conrft, hu2024flare}, safety constraints~\cite{zhang2025safevla}, reasoning consistency~\cite{kim2025robot}, and data-selection preferences~\cite{zhang2025reinbot, chen2025tgrpo}. In addition, RL can produce domain specialists that surpass expert policies; these specialists can be leveraged to enhance the policies of general-purpose VLA models~\cite{xu2024rldg}. Representative RL-based post-training methods are summarized in Table~\ref{table:Muti}.

\subsubsection{Visual Interaction Prediction}

\label{Sec. VIP}

Visual interaction prediction involves forecasting the interactions between the embodiment and environment under actions. This capability requires generating action sequences, simulating the resulting embodiment behavior, and predicting environmental changes. Thus, unlike the solely kinematics learning discussed in Sec.~\ref{Sec:Forward kinematics learning}, these methods are categorized as multi-component integration. The approach is first shown to be effective for pre-training on unlabeled demonstration data~\cite{wu2023unleashing, cheang2024gr} and has since become a widely adopted post-training strategy for VLA models.

One line of work forecasts future observation frames using action-conditioned predictive models. Frame prediction can be embedded in model predictive control (MPC) for forward-kinematics learning, where control objectives are optimized by minimizing pixel-wise differences between predicted and target frames~\cite{zhao2024vlmpc, chen2025vision}. Predictive models also facilitate inverse-kinematics learning: Zhou et al.~\cite{zhou2024minedreamer} generate future frames from current observations and instruction embeddings, then use both predicted and actual frames to infer actions. Beyond single-frame prediction, interaction trajectories can be forecast directly~\cite{shridhar2025generative, li2025hamster}. Alternatively, forecasting can occur in a latent feature space by predicting future observation representations~\cite{hu2024video, zhang2025up, chen2024moto, zheng2025flare}. Latent space prediction reduces the complexity of pixel-level synthesis and enables robust dynamics modeling even when frame predictions are imperfect.

Beyond frame forecasting, interaction prediction can be formulated via various strategies~\cite{zeng2024learning, li2024virt, jin2024learning, hao2024language}. Zeng et al.~\cite{zeng2024learning} predict intermediate interaction frames between action initiation points and subsequently apply a detection module to localize interaction regions. Li et al.~\cite{li2024virt} generate action guidance by masking target objects and constructing point-flow representations, rather than synthesizing full-frame predictions. Ni et al.~\cite{ni2024generate} incorporate CoT to predict the next sub-task target and use it to guide the policy. Although these directions remain relative to direct frame forecasting, they offer promising avenues for advancing interaction prediction. Representative studies are summarized in Table~\ref{table:Muti}.

\subsubsection{Active Dataset Processing}

\label{Sec. EFT}

Manipulation datasets for VLA models are costly to construct, motivating the use of minimal yet informative demonstration sets for post-training. Achieving this efficiency requires for active selection and augmentation strategies that leverage knowledge of tasks, environments, and embodiments.

An intuitive approach to active processing is selecting informative demonstrations for post-training, as demonstration quality also determines VLA policy gains. Lin et al.~\cite{lin2024data} show that diversity over environments and objects outweighs raw demonstration count. Accordingly, informativeness-based selection and weighting have been proposed: Bagatella et al.~\cite{bagatella2024active} estimate information gain to curate fine-tuning datasets, while Wu et al.~\cite{wu2024swbt} select high-similarity imperfect trials to enhance data efficiency. Demonstration importance also guides collection policies: Zheng et al.~\cite{zheng2025dataplatter} prioritize spatial reasoning scenarios, and Gao et al.~\cite{gao2024efficient} decompose tasks by object properties to reduce samples via combinatorial data designs.

Limited high-cost demonstrations can be augmented with low-cost, automatically collected or generated samples~\cite{zhouautonomous, jin2024learning}. Self-improvement methods use foundation model guidance for autonomous demonstration gathering, reducing reliance on expert data~\cite{zhouautonomous}. Generative frameworks synthesize manipulation trajectories to expand datasets~\cite{jin2024learning}. Post-training efficiency can be further improved through curriculum design and augmentation~\cite{wen2025dexvla, wang2025robobert}; examples include decomposing long-horizon tasks into shorter sub-tasks for curriculum learning~\cite{wen2025dexvla} and applying data augmentation to diversify inputs~\cite{wang2025robobert}. Representative active dataset processing methods are summarized in Table~\ref{table:Muti}.

\begin{table*}[!htb]\scriptsize
\centering
\caption{Performance Comparison on the LIBERO Benchmark}
\begin{threeparttable}
\label{tab:vla_libero}

\begin{tabular}
{
  >{\centering\arraybackslash}m{1cm}
  >{\centering\arraybackslash}m{4cm}
  >{\centering\arraybackslash}m{3.5cm}
  >{\centering\arraybackslash}m{0.65cm}
  >{\centering\arraybackslash}m{0.65cm}
  >{\centering\arraybackslash}m{0.65cm}
  >{\centering\arraybackslash}m{0.65cm}
  >{\centering\arraybackslash}m{0.65cm}
  >{\centering\arraybackslash}m{1.5cm}
}
\toprule
Date & Backbone & Method & Spatial & Object & Goal & Long & Avg. & Class \\
\midrule
2024.05 & Light-CNN + T5 & Octo~\cite{team2024octo} & 78.9 & 85.7 & 84.6 & 51.1 & 75.1 & FM \\
2024.06 & Llama 2 7B + SigLIP + DINOv2 & OpenVLA~\cite{kimopenvla} & 84.7 & 88.4 & 79.2 & 53.7 & 76.5 & FM \\
2024.10 & PaliGemma-3B & $\pi_0$~\cite{intelligence2025pi_} & 96.8 & 98.8 & 95.8 & 85.2 & 94.2 & FM \\
2024.11 & CLIP & CLIP-RT~\cite{kang2024clip} & 95.2 & 99.2 & 94.2 & 83.8 & 93.1 & HTM \\
2024.11 & Prismatic & CogACT~\cite{li2024cogact} & 97.2 & 98.0 & 90.2 & 88.8 & 93.6 & FM \\
2024.12 & DiVLA & CoA-VLA~\cite{li2024improving} & 85.3 & 93.1 & 85.8 & 55.0 & 79.8 & ADP \\
2024.12 & Phi-3-Vision & TraceVLA~\cite{zheng2024tracevla} & 84.6 & 85.2 & 75.1 & 54.1 & 74.8 & ERM \\
2025.01 & miniVLA-VQ & InSpire~\cite{zhang2025inspire} & 13.7$^*$ & 20.1$^*$ & 12.7$^*$ & 8.0$^*$ & 13.6$^*$ & EEM \\
2025.01 & PaliGemma 3B & SpatialVLA~\cite{qu2025spatialvla} & 88.2 & 89.9 & 78.6 & 55.5 & 78.1 & ERM \\
2025.01 & pi0 & $\pi_0$-Fast~\cite{pertsch2025fast} & 96.4 & 96.8 & 88.6 & 60.2 & 85.5 & AHD \\
2025.02 & Qwen2-VL + ScaleDP & DexVLA~\cite{wen2025dexvla} & 97.2 & 99.1 & 95.6 & -- & -- & ADP \& HTM \\
2025.02 & OpenVLA & OpenVLA-OFT~\cite{kim2025fine} & 95.2 & 94.2 & 95.2 & 93.2 & 94.5 & AHD \& ERM \\
2025.02 & OpenVLA & VLA-Cache~\cite{xu2025vla} & 83.8 & 85.8 & 76.4 & 52.8 & 74.7 & ERM \\
2025.03 & VILA-U & CoT-VLA~\cite{zhao2025cot} & 87.5 & 91.6 & 87.6 & 69.0 & 81.1 & VIP \\
2025.03 & DiT + CLIP + DINOv2 & Dita~\cite{hou2025dita} & 97.4 & 94.8 & 93.2 & 83.6 & 92.3 & AHD \\
2025.03 & Temporal Transformer + ViT & OMLA~\cite{zhu2025efficient} & 0.86 & 0.69 & 0.58 & -- & -- & AHD \\
2025.03 & CLIP + SigLIP & OTTER~\cite{huang2025otter} & 84.0 & 89.0 & 82.0 & -- & 85 & ERM \\
2025.03 & Eagle-2 + DiffusionExpert & GR00T N1~\cite{bjorck2025gr00t} & 94.4 & 97.6 & 93.0 & 90.6 & 93.9 & FM \\
2025.05 & Llama 2 7B + VIT + PointNet & 3D-CAVLA~\cite{bhat20253d} & 98.2 & 99.8 & 98.2 & 96.1 & 98.1 & ERM \\
2025.05 & InternLM2 1.8B + SigLIP + DINOv2 & GraspVLA~\cite{deng2025graspvla} & 82.0$^*$ & 91.2$^*$ & 94.1$^*$ & -- & -- & FM \\
2025.05 & OpenVLA-OFT & RIPT-VLA~\cite{tan2025interactive} & 92.7 & 95.6 & 98.4 & 87.5 & 93.6 & RL \\
2025.05 & OpenVLA & VLA-RL~\cite{lu2025vla} & 90.2 & 91.8 & 82.2 & 59.8 & 81.0 & RL \\
2025.06 & Prismatic-7B & CronusVLA~\cite{li2025cronusvla} & 90.1 & 94.7 & 91.3 & 68.7 & 86.2 & ERM \& AHD \\
2025.06 & ECoT & Fast ECoT~\cite{duan2025fast} & 83.0 & 85.0 & 83.0 & 69.0 & 80.0 & HTM \\
2025.06 & SmolVLM 2.25B & SmolVLA~\cite{shukor2025smolvla} & 93.0 & 94.0 & 91.0 & 77.0 & 88.8 & FM \\
2025.06 & OpenVLA & TGRPO~\cite{chen2025tgrpo} & -- & 91.0 & -- & -- & -- & RL \\
2025.06 & Action World Model & WorldVLA~\cite{cen2025worldvla} & 87.6 & 96.2 & 83.4 & 60.0 & 81.8 & VIP \\
2025.07 & LLaMA2-7B & VOTE~\cite{lin2025vote} & 98.8 & 99.8 & 97.6 & 95.6 & 98.0 & AHD \\
2025.07 & OpenVLA & VQ-VLA~\cite{wang2025vq} & -- & -- & 72.4 & -- & -- & AHD \\
2025.08 & Prismatic-7B + SigLIP + DINOv2 & Discrete Diffusion VLA~\cite{liang2025discrete} & 97.2 & 98.6 & 97.4 & 92.0 & 96.3 & AHD \\
2025.08 & Primatic & GeoVLA~\cite{sun2025geovla} & 98.4 & 99.0 & 96.6 & 96.6 & 97.7 & ERM \\
2025.08 & Prismatic-7B + SigLIP + DINOv2 & MemoryVLA~\cite{shi2025memoryvla} & 98.4 & 98.4 & 96.4 & 93.4 & 96.7 & ERM \\
\bottomrule
\end{tabular}
\begin{tablenotes}
    \item[1] $*$: results on unseen tasks.
    \item[2] Abbreviations: (1) Foundation Model: FM; (2) Hierarchical Task Manipulation: HTM; (3) Active Dataset Processing: ADP; (4) Enhanced Representation for Manipulation: ERM; (5) Reinforcement Learning: RL.
\end{tablenotes}
\end{threeparttable}
\end{table*}

\begin{table*}[!htb]\scriptsize
\centering
\caption{Performance Comparison on the Calvin benchmark}
\label{tab:Calvin}
\begin{threeparttable}

\begin{tabular}
{
  >{\centering\arraybackslash}m{1cm}
  >{\centering\arraybackslash}m{3.5cm}
  >{\centering\arraybackslash}m{3cm}
  >{\centering\arraybackslash}m{0.65cm}
  >{\centering\arraybackslash}m{0.65cm}
  >{\centering\arraybackslash}m{0.65cm}
  >{\centering\arraybackslash}m{0.65cm}
  >{\centering\arraybackslash}m{0.65cm}
  >{\centering\arraybackslash}m{1cm}
  >{\centering\arraybackslash}m{1.5cm}
}
\toprule
Date & Backbone & Method & 1 & 2 & 3 & 4 & 5 & Avg. Len & Class \\
\midrule
2024.06 & Transformer & DISCO~\cite{hao2024language} & 94.7 & 82.9 & 71.0 & 58.8 & 49.4 & 3.57 & VIP \\
2024.12 & GPT + T5 + ViT & Moto~\cite{chen2024moto} & 89.7 & 72.9 & 60.1 & 48.4 & 38.6 & 3.10 & VIP \\
2024.12 & Stable Video Diffusion & VPP~\cite{hu2024video} & 96.5 & 90.9 & 86.6 & 82.0 & 76.9 & 4.33 & VIP \\
2025.01 & Phi-1.5 + ViT & UP-VLA~\cite{zhang2025up} & 92.8 & 86.5 & 81.5 & 76.9 & 69.9 & 4.08 & VIP \\
2025.02 & ViT + BERT + OpenFlamingo & RoboBERT~\cite{wang2025robobert} & 95.3 & 85.7 & 75.4 & 66.3 & 56.2 & 3.79 & ADP \& EEM \\
2025.03 & DiT + CLIP + DINOv2 & Dita~\cite{hou2025dita} & 94.5 & 82.5 & 72.8 & 61.3 & 50.0 & 3.61 & AHD \\
2025.03 & Roboflamingo & RoboFlamingo-Plus~\cite{li2025roboflamingo} & 90.0 & 73.0 & 57.5 & 46.0 & 31.5 & 3.00 & ERM \\
2025.05 & Seer & RFTF~\cite{shu2025rftf} & 96.4 & 91.7 & 86.7 & 80.7 & 74.1 & 4.30 & RL \\
2025.07 & GPT-2 + ViT & DreamVLA~\cite{zhang2025dreamvla} & 98.2 & 94.6 & 89.5 & 83.4 & 78.1 & 4.44 & VIP \\
\bottomrule
\end{tabular}
\begin{tablenotes}
    \item[1] Abbreviations: (1) Visual Interaction Prediction: VIP; (2) Active Dataset Processing: ADP; (3) Action Head Designing: AHD; (4) Enhanced Representation for Manipulation: ERM; (5) Reinforcement Learning: RL.
\end{tablenotes}
\end{threeparttable}
\end{table*}

\section{Comparison on Standard Benchmarks}

\subsection{Benchmark Description}

LIBERO~\cite{liu2023libero} is a benchmark for multi-task and lifelong robot learning, comprising 130 manipulation tasks grouped into four suites: \textbf{LIBERO-Spatial:} evaluates spatial reasoning under perceptual ambiguity; \textbf{LIBERO-Object:} tests continual generalization to novel object categories; \textbf{LIBERO-Goal:} measures goal conditioning with fixed objects and layouts; \textbf{LIBERO-100:} assesses efficiency and compositional generalization across short- (LIBERO-90) and long-horizon (LIBERO-LONG) tasks. Success rate is the evaluation metric.

CALVIN~\cite{mees2022calvin} is an open-source benchmark for long-horizon, language-conditioned robotic manipulation. Agents interpret and execute sequences of unconstrained natural language commands over extended interactions. Tasks comprise 1-5 sub-tasks. Training data cover sets A, B, and C, while set D is held out for testing, enabling rigorous generalization evaluation. The benchmark also supports flexible sensor-suite configurations to accommodate diverse perception setups. Evaluation reports the average successful sequence length and success rates stratified by sub-task count.

\subsection{Results Analysis on LIBERO Benchmark}

Table~\ref{tab:vla_libero} reports LIBERO~\cite{liu2023libero} performance from a range of algorithms. For completeness, the table includes both post-training methods and widely used foundation models. Over the past sixteen months, the average success rate has risen from 75\% to 98\%. At the task-set level, aside from the long-horizon suite, SOTA success rates on the remaining suites exceed 98\%. Consequently, meaningful headroom now primarily lies in improving long-horizon manipulation.

A closer examination of post-training design indicates that action-head designing yields obviously and ``plug-and-play” gains. For example, OpenVLA-OFT~\cite{kim2025fine}, built on OpenVLA~\cite{kimopenvla}, improves the average success rate from 76.5\% to 94.5\% by adopting parallel decoding, action chunking, and continuous outputs. Similar observations appear across other works for action head designing~\cite{lin2025vote, hou2025dita, kim2025fine}.

Another effective strategy is to explicitly leverage additional perception modalities or temporal memory to enrich manipulation-specific representations. 3D-CAVLA~\cite{bhat20253d} attains the best reported result (98.1\%) by exploiting point cloud geometry. MemoryVLA~\cite{shi2025memoryvla} builds a memory buffer over historical features to improve temporal understanding, achieving 96.7\%. These findings suggest that, while large-scale pretraining confers broadly generalized vision–language features, manipulation-oriented representations remain underexplored and offer additional advantages.

Finally, although RL-based often surpasses purely supervised learning on other tasks; however, RL-tuned VLA models still trail the best post-training SOTA in several cases. Early gains from hierarchical policy formulations (e.g., CLIP-RT~\cite{kang2024clip}) have been less prominent in recent leaderboards, potentially because the current benchmark difficulty underemphasizes structures where hierarchical methods most clearly excel, especially in truly long-horizon settings.

\subsection{Results Analysis on CALVIN Benchmark}

Experimental results on the CALVIN benchmark show rapid progress: the average successful sequence length increased from roughly 3.5~\cite{hao2024language} to 4.3-4.4 tasks~\cite{zhang2025dreamvla} (Table~\ref{tab:Calvin}). A closer check by instruction length indicates that gains are pronounced at longer horizons: success on length-1 tasks rose from 89.7\% to 98.2\%, whereas length-5 tasks improved from 31.5\% to 78.1\%. Despite these gains, performance on long-horizon tasks remains the primary bottleneck within this benchmark.

A related observation is that simply adding auxiliary perceptual inputs to enhance representations for manipulation does not yield substantial improvements on long-horizon tasks. For example, RoboFlamingo-Plus~\cite{li2025roboflamingo}, despite incorporating depth information, exhibited limited gains. This differs from findings on the LIBERO~\cite{liu2023libero} and likely reflects CALVIN’s emphasis on long-horizon sequencing. Representation-only enhancements may assist low-level manipulation but provide limited benefit for long-horizon planning.

Within this benchmark, a common direction is predictive control via visual interaction prediction: methods such as DreamVLA~\cite{zhang2025dreamvla} and VPP~\cite{hu2024video} leverage interaction frame prediction to construct task-relevant latent dynamics, a design that plausibly benefits long-horizon performance. In addition, RL-based post-training has demonstrated strong results on CALVIN~\cite{shu2025rftf}. Although RL post-training for VLA remains underexplored, its potential is evident.

\subsection{Takeaways}

After comparing the experimental outcomes from the LIBERO~\cite{liu2023libero} and CALVIN~\cite{mees2022calvin} benchmarks, the following technical observations are summarized.

\begin{itemize}
    \item Carefully designed post-training, including action-head design and objective selection, can substantially increase backbone VLA model performance on downstream tasks.
    \item Enhanced representations (e.g., 3D spatial or tactile cues) can improve short-horizon manipulation; however, gains on long-horizon tasks may be limited unless coupled with techniques, such as world modeling and hierarchical control, to curb error accumulation.
    \item Training VLA models to predict future observations or latent dynamics benefits long-horizon control and can be readily integrated with existing VLA models and trained with large datasets.
    \item RL and hierarchical task manipulation remain under-explored yet promising for VLA post-training. A minimal pipeline is recommended: a language/visual planning head at the high level; a VLA executor at the low level; offline BC/RLPD on demonstrations and replay, followed by online RL with rewards and safety constraints.
    \item Benchmark design should emphasize more on diversity and difficulty. Beyond higher task complexity (where LIBERO average success has approached $98\%$), future suites can target longer horizons, partial observability, and compositional generalization to provide more comprehensive evaluation.
\end{itemize}

\section{Challenges and Future Trends}

\subsection{Benchmark and Evaluation}

\subsubsection{Standardized Benchmarks and Statistical Protocols}

Despite numerous techniques for VLA models, fair cross-study comparison remains a challenge. Existing simulator-based benchmarks (e.g., LIBERO~\cite{liu2023libero}, CALVIN~\cite{mees2022calvin}) mitigate only part of the variance because experimental settings such as hyperparameters and environment randomization are different across studies.
This stochasticity from random seeds can introduce larger variance in experimental results; for instance, success-rate fluctuations above 30\%  have been reported~\cite{zhang2024effective}. While simulators can hold seeds and initial states fixed, real-world evaluations rarely afford the same level of control, complicating reproducibility. Hence, statistical power is essential, yet the choice of an adequate number of repetitions remains an open question in robotics evaluation~\cite{andrychowicz2021matters}.

Community efforts have moved toward standardized task suites in simulation~\cite{liu2023libero,mees2022calvin}. But there is still a need for a unified evaluation benchmark in robotic learning, analogous to ImageNet in CV~\cite{deng2009imagenet}. The emergence of ImageNet catalyzes numerous innovations and directions within CV. Lessons drawn from ImageNet suggest that a widely adopted standard could accelerate progress in the VLA model community.

Human motor evaluation relies on standardized scales and specified testing procedures to improve objectivity in assessment. At the other end of the spectrum, elite competitions (e.g., large international sporting events) enforce strict rules and controlled conditions to measure peak performance consistently across participants and venues. These approaches illustrate how prescriptive protocols and transparent scoring can reduce evaluator bias and foster reliable comparisons. Although primarily evaluative, their structure and controlled variability parallel principles of structured, variable practice that support comprehensive motor-skill development (Sec.~\ref{Sec: human comprehensive improvement}). Lessons from standardized human assessments and the ``ImageNet effect'' in computer vision~\cite{deng2009imagenet} motivate a two-track protocol for VLA: (i) a simulation track with fixed task splits, pre-defined initialization distributions, seed control, uniform evaluation criteria, and a statistical analysis. (ii) a real-robot track hosted under shared conditions (e.g., community workshop or conference) that specifies protocols, such as hardware profiles and environmental settings.

\subsubsection{Diversity of Evaluation Metrics}

Prevailing evaluations of VLA models rely predominantly on task success rate, a coarse indicator that may obscures some critical factors such as real-time latency, adherence to safety constraints, motion smoothness and stability, calibration fidelity, and long-horizon reliability. This narrow emphasis hinders fair comparison across methods and can misrepresent practical capability. Recent work has begun to broaden the metric space; for example, Zhang et al.~\cite{zhang2025safevla} incorporate safe reinforcement learning into VLA post-training. Nevertheless, systematic evaluation frameworks remain limited.

By contrast, human motor assessment employs standardized scales and kinematic analyses that extend beyond binary pass/fail, emphasizing dimensions such as smoothness and fluency, reaction time, and temporal consistency over extended activities. These protocols demonstrate how prescriptive scoring yields a faithful characterization of capability. Within VLA post-training, such measures can be directly adopted or lightly adapted to evaluate both outcome and process quality of behavior. Moreover, these indices can inform the design of robot learning objectives and constraints, aligning optimization with safety and motion-quality considerations. Because evaluation spans perception, embodiment, and task understanding, efforts from this perspective do not fit neatly into a single group within the proposed taxonomy; metric definitions can be specified per axis as needed to enable comprehensive assessment.

\subsubsection{Structured and Effective Dataset}

Although VLA models benefit from large-scale data, collecting robot manipulation datasets is costly, yielding markedly fewer samples than in other domains. Even after aggregating multiple sources, the overall scale remains limited~\cite{o2024open}. Beyond quantity, downstream performance depends critically on data quality. The field therefore faces a dual requirement: increase volume while maintaining high-quality manipulation data.

Recent efforts aim to expand scale by leveraging external sources~\cite{cheang2024gr} and by generating synthetic manipulation trajectories~\cite{bjorck2025gr00t}. Complementary initiatives target quality through standardized collection protocols~\cite{o2024open}, higher-fidelity simulators~\cite{makoviychuk2isaac}, data-quality assessment~\cite{bagatella2024active}, and acquisition policies that prioritize informative samples~\cite{gao2024efficient}. Despite these advances, a data–performance gap persists.

Human motor learning achieves high sample efficiency from comparatively few demonstrations, yet it further benefited from the quality and structure of instruction. Established strategies, such as curriculum design and expert coaching, accelerate skill acquisition and robustness~\cite{czyz2024high}. Within the proposed taxonomy, these strategies span multiple axes: high-quality, structured instruction aligns with comprehensive improvement (Sec.~\ref{Sec: human comprehensive improvement}), whereas expert coaching and curricula primarily support deepening task comprehension (Sec.~\ref{Sec: human Deepening Task Comprehension}). These insights motivate post-training that exploit quality-weighted and structured data, including: (i) curriculum scheduling that stages tasks by increasing complexity;
(ii) teacher–student distillation and preference-guided objectives that prioritize high-quality demonstrations;
(iii) active selection of informative demonstrations.

\subsection{Manipulation Ability}

\subsubsection{Multimodal Perception Fusion}

Current VLA models inherit strong vision-language representations from pre-trained encoders. However, most pre-training pipelines omit other critical modalities for manipulation, including tactile/force signals, depth/point clouds, and proprioception. This bias can yield policies that generalize visually yet are poor in contact-rich settings or when visual inputs are degraded during control.

Recent studies have begun to incorporate additional modalities into VLA architectures. Examples include vision–tactile–language–action models~\cite{zhang2025vtla} and VLA variants augmented with force feedback~\cite{yu2025forcevla}, which report gains on contact-intensive tasks. Despite these advances, multimodal data acquisition and efficient fusion within existing VLA frameworks remain open challenges.

Human motor execution relies on tightly integrated multimodal sensing: vision, haptics, proprioception, and audition. Through sensory reweighting, when one channel is unreliable, humans can maintain performance (e.g., typing without gaze) by leveraging procedural memory, anticipatory feedforward control, and rapid error correction. These mechanisms align with the strategies for enhancing environmental perception (Sec.~\ref{Sec: human enhance environmental perception}) and motivate robotic perception, especially under uncertainty. Insights from human multimodal integration suggest post-training protocols that: (i) introduce additional modalities (tactile, proprioception); (ii) apply uncertainty-aware fusion to mimic human sensory reweighting; (iii) train with modality dropout/occlusion to harden policies against sensor failures and improve reliability.

\subsubsection{Long-Horizon, Hierarchical Skill Composition}

On long-horizon tasks, VLA policies often degrade relative to short-horizon manipulation due to accumulated prediction errors. End-to-end architectures typically lack explicit self-correction mechanisms, allowing early mispredictions to propagate into late-stage failures. Moreover, data requirements grow with task length, while demonstration collection becomes increasingly costly.

Recent approaches introduce temporal structure assisting action execution: task planners decompose goals into sub-goals; hierarchical policies and skill libraries execute reusable subroutines~\cite{kang2024clip}; and progress monitors provide mid-trajectory feedback and recovery triggers~\cite{lee2024affordance}. Despite these advances, robust task planning and reliable error correction remain open challenges for long-horizon VLA models.

The hierarchical control in robot is inspired by human motor control: high-level circuits plan and sequence actions while low-level circuits execute and refine motor commands~\cite{merel2019hierarchical}. Skill chunking, procedural memory, and feedback control support extended routines, with online corrections driven by sensory prediction errors. Such mechanisms limit error accumulation, reduce cognitive load, and maintain fluency across extended behaviors. This part aligns with the strategy for deepening task comprehension (Sec.~\ref{Sec:Deepening Task Comprehension}). These observations motivate post-training protocols that: (i) learn and distill reusable skill primitives; (ii) incorporate progress monitors and uncertainty triggers for self-correction; (iii) decompose long tasks into sub-goals with measurable progress signals.

\subsubsection{Real-time Response and Execution}

Large VLM provides strong perceptual capabilities for VLA systems, but introduces computational latency and variance, impairing real-time robotic manipulation. During deployment, controllers require low and predictable end-to-end delay; otherwise, stability and responsiveness degrade, particularly in contact-rich tasks and dynamic scenes.

Recent efforts address real-time constraints by reducing decision latency and compute: action chunking and streaming/parallel decoding shorten per-step inference~\cite{bjorck2025gr00t, song2025accelerating}, while quantization, pruning, distillation, and mixture-of-experts with sparse routing accelerate vision-language feature extraction~\cite{song2025ceed}. These techniques improve efficiency but may incur accuracy degradation, necessitating careful calibration and robustness checks.

Except for the high-efficiency nature of bio-intelligence itself, human motor control achieves fast responses through forward internal models that anticipate body-environment interactions, complemented by reflexive feedback loops that correct residual errors at short latencies. A library of procedural motor primitives further enables rapid execution in familiar contexts. These mechanisms align with the proposed taxonomy: the construction and reuse of efficient motor primitives support deepening task comprehension  (Sec.~\ref{Sec: human Deepening Task Comprehension}), while multi-component integration that predicts interaction contributes to comprehensive improvement (Sec.~\ref{Sec: human comprehensive improvement}).

Insights from human control motivate the post-training process that: (i) distill large perception backbones into compact modules and construct reusable skill primitives, with regularization for stable chunked execution. (ii) add a predictive head that predicts actions under sensor and compute delays. (iii) deploy high-efficiency implementations of encoders (e.g., event-driven or neuromorphic substrates).

\subsection{Knowledge Transfer}

\subsubsection{Continual/Lifelong Skill Transfer}

Pre-/post-training pipelines leverage priors in large backbones; however, empirical results indicate that fully unfreezing the vision–language encoder during post-training is often required for strong downstream performance~\cite{wang2025robobert,zhang2024effective}. This full-parameter fine-tuning risks catastrophic forgetting of previously acquired representations and manipulation skills.

To acquire new skills without erasing prior ones, recent work integrates continual-learning mechanisms into VLA model post-training~\cite{liang2024never}. Typical strategies include knowledge distillation from teacher checkpoints and curated replay of prior trajectories. Despite progress, many evaluations rely on limited skill suites and resemble multi-task training rather than lifelong learning.

Human motor learning mitigates forgetting through multi-timescale adaptation (fast, explicit updates with slow, implicit consolidation), contextualization of control policies, and reuse of procedural primitives. Sleep-associated replay strengthens recently learned routines, while context-dependent retrieval reduces interference among overlapping skills. Compositional reuse of motor primitives enables rapid acquisition of novel behaviors with minimal degradation of previously mastered tasks. These mechanisms align with the proposed taxonomy in the group of comprehensive improvement (Sec.~\ref{Sec: human comprehensive improvement}).

Insights from human learning motivate post-training protocols that: (i) distill reusable skill primitives and enforce subgoal-aware objectives to stabilize composition; (ii) couple fast task-specific heads with slow-consolidation mechanisms; (iii) incorporate uncertainty-triggered context gating for policy selection and interference reduction.

\subsubsection{Cross-embodiment Transfer}

Scarcity of manipulation data encourages pre-training on heterogeneous robot platforms and internet demonstrations, creating an embodiment mismatch between pre-training sources and deployment targets. Such gaps pose a cross-embodiment transfer problem.

Recent studies pursue two complementary directions: (i) embodiment-conditioned adaptation (e.g., embodiment-specific adapters or conditioning tokens) that retain robust vision–language perception while adapting control~\cite{brohan2023rt}; and (ii) embodiment-agnostic priors (e.g., affordance- or goal-centric descriptors) integrated into VLA models to accelerate transfer~\cite{nasirianyrt}. Despite progress, reliable zero-shot re-targetting and safety under morphology shifts remain open challenges.

Humans achieve cross-embodiment transfer via effector-independent motor representations, observational learning from demonstrators with different morphologies, and rapid binding of skills to new effectors. Dual explicit–implicit learning processes, combined with closed-loop feedback control, support swift adjustment of motor commands to novel embodiments. These mechanisms align with the proposed taxonomy: learning from demonstration deepens task comprehension (Sec.~\ref{Sec: human Deepening Task Comprehension}); and dual-process learning contributes to comprehensive improvement (Sec.~\ref{Sec: human comprehensive improvement}).

These insights motivate post-training protocols that: (i) construct task representations within VLA models; (ii) design dual-process learning modules that mirror explicit-implicit learning policy in humans; (iii) incorporate closed-loop adaptation for fast online tuning under embodiment mismatch.

\subsubsection{Cross-environment Transfer}

Most VLA systems are evaluated in laboratory settings or specialized simulators, where SOTA methods report high task success rates and some degree of generalization~\cite{intelligence2025pi_, intelligence2025pi05}. Nevertheless, environment-level generalization remains limited. Existing evaluations and post-training protocols predominantly optimize for task-level performance within fixed or narrowly perturbed environments, leaving cross-environment deployment a persistent challenge.

Humans execute motor skills reliably despite environmental changes by combining rapid contextual inference with memory expression switching. Before and during action, the nervous system infers the current context, retrieves appropriate motor schemas from prior experience, and adaptively reconfigures which representations are expressed, rather than globally rewriting the underlying ``weights." This representational re-routing increases robustness to environmental perturbations. Beyond intrinsic mechanisms, practicing across diverse real settings and high-fidelity virtual scenarios further builds invariances and accelerates on-the-fly adaptation. These mechanisms map to multiple axes in the proposed taxonomy: contextual inference aligns with enhancing environmental perception (Sec.~\ref{Sec: human enhance environmental perception}); schema retrieval and representational routing support deepening task comprehension (Sec.~\ref{Sec: human Deepening Task Comprehension}); and structured, variable practice contributes to comprehensive improvement (Sec.~\ref{Sec: human comprehensive improvement}).

These observations suggest that VLA post-training can be improved through: (i) learn to infer environment context and route behavior through environment-conditioned memory modules; (ii) disentangle skill representations from environment factors; (iii) incorporate diverse real and simulated environments into training data.

\subsection{Safety and Ethics}

\subsubsection{Explainability for VLA Models}

Despite its importance, XAI for VLA remains underexplored. Existing efforts largely emphasize error identification and recovery during manipulation, as well as learning and exposing manipulation pattern structure. Because explainability is ultimately judged by human understanding, principles from human motor learning offer a structured pathway to mechanism-level interpretability. Incorporating constructs such as memory, hierarchical control, and manipulation primitives into model design can yield interfaces and behaviors that are easier to analyze, validate, and trust, thereby contributing to improving explainability for VLA Models. This theme does not map onto a single axis of the proposed taxonomy; rather, it is cross-cutting, and insights from human motor mechanisms can enhance XAI from multiple angles.

\subsubsection{Safety in VLA Models}

Because VLA systems will be deployed in the physical world, unsafe actions can cause harm to people, equipment, and surroundings. Predominantly learning-based VLA models are data-driven, making behavior under distribution shift and rare corner cases difficult to predict. Consequently, safety must be guaranteed not only during training but also for deployment.

Safety during learning is often promoted by human-in-the-loop protocols that provide corrective feedback or guard exploration~\cite{chen2025conrft}, and by optimizing objectives that include safety preferences or soft constraints. However, these approaches primarily encode safety preferences into the policy and typically lack end-to-end guarantees; protection will degrade when the system operates autonomously in open-world conditions.

In human motor learning, safety is maintained through strong priors about hazards and adherence to external rules and procedures. Critically, demonstrations and practice are curated so that exemplars themselves respect safety constraints, reinforcing robust habits and excluding unsafe behaviors. These strategies align with the deepening task comprehension within the proposed taxonomy (Sec.~\ref{Sec: human Deepening Task Comprehension}). Insights from humans suggest post-training should explicitly: (i) impose hard safety constraints that cannot be overridden; (ii) ensure demonstrations and preference data comply with constraints; (iii) integrate runtime monitoring and formal checks for critical invariants.

\section{Conclusion}

This survey presents post-training techniques for VLA models in robotic manipulation through the lens of human motor learning based on Newell’s constraints-led theory, organized around three core components: environments, embodiments, and tasks. A high-level taxonomy, mirroring the way of human skill acquisition, categorizes research efforts aimed at enhancing model capabilities across these components, with representative methods summarized within each category. Challenges and limitations of current VLA post-training approaches are then identified, followed by a summary of human strategies to address them. By aligning VLA post-training with human learning processes, this review offers a novel perspective for understanding and improving model adaptation. However, as a review study, this paper does not delve deeply into technical implementation details. Instead, we aim to provide a practical reference for future researchers to investigate VLA post-training or to inform model design from the perspective of human learning.

{\small
\bibliographystyle{IEEEtran}
\bibliography{mybib}
}

\end{document}